\documentclass{article}
\usepackage[preprint, nonatbib]{neurips_2024}
\usepackage[pagebackref,breaklinks,colorlinks,citecolor=blue]{hyperref}
\usepackage{float}
\usepackage{url}
\usepackage{microtype}
\usepackage{graphicx}
\usepackage{threeparttable}
\usepackage{subcaption}
\usepackage{booktabs} 
\usepackage{multirow}
\usepackage{comment}
\usepackage{tabularx}
\usepackage{xcolor}
\usepackage{makecell}
\usepackage{wrapfig}
\usepackage{colortbl}
\usepackage{enumitem}

\def\btheta{\boldsymbol\theta}
\def\bx{\mathbf{x}}
\def\D{\mathcal{D}}

\newcommand{\heading}[1]{\noindent\textbf{#1}}

\newcommand{\method}{Bella\xspace}

\newcommand{\hatff}{\hat{\boldsymbol{\phi}}}
\definecolor{grey}{rgb}{0.9,0.9,0.9} 
\definecolor{white}{rgb}{1,1,1} 

\newcommand{\dtoprule}{\specialrule{1pt}{0pt}{\belowrulesep}
            }
\newcommand{\dbottomrule}{
            \specialrule{1pt}{0pt}{\belowrulesep}%
            }

\newcommand{\nlr}{r} 

\newcommand{\A}{\mathbf{A}}

\newcommand{\B}{\mathbf{B}}

\RequirePackage{xspace}
\makeatletter
\DeclareRobustCommand\onedot{\futurelet\@let@token\@onedot}
\def\@onedot{\ifx\@let@token.\else.\null\fi\xspace}

\def\eg{\emph{e.g}\onedot} 

\def\ie{\emph{i.e}\onedot}

\makeatother

\usepackage{amsmath}
\usepackage{amssymb}
\usepackage{mathtools}
\usepackage[capitalize,noabbrev]{cleveref}
\usepackage{xspace}
\usepackage{booktabs}

\usepackage{tcolorbox}
\newcommand{\rqq}[1]{
\begin{center}
    \begin{tcolorbox}[width=\columnwidth, colback=orange!10!white, colframe=white!15,left=2pt,right=2pt,top=1pt,bottom=1pt,arc=4pt,auto outer arc]
    \textit{#1}
    \end{tcolorbox}
\end{center}
}

\title{Bayesian Low-Rank Learning (Bella): \\A Practical Approach to Bayesian Deep Learning}

\author{Bao Gia Doan$^{1}$$^\ast$, Afshar Shamsi$^{2}$$^\ast$, Xiao-Yu Guo$^{1}$, Arash Mohammadi$^{2}$\\ \textbf{Hamid Alinejad-Rokny}$^{3}$ \textbf{Dino Sejdinovic}$^{1}$, \textbf{Damien Teney}$^4$\\ \textbf{Damith C. Ranasinghe}$^{1}$, \textbf{Ehsan Abbasnejad}$^{1}$\\
$^{1}$The University of Adelaide, Australia \quad $^{2}$Concordia University, Canada \\
$^{3}$ UNSW Sydney, Australia \quad $^4$Idiap Research Institute, Switzerland}

\begin{document}

\maketitle

\begin{abstract}

Computational complexity of Bayesian learning is impeding its adoption in practical, large-scale tasks. Despite demonstrations of significant merits such as improved robustness and resilience to unseen or out-of-distribution inputs over their non-Bayesian counterparts, their practical use has faded to near insignificance. In this study, we introduce an innovative framework to mitigate the computational burden of Bayesian neural networks (BNNs).  
Our approach follows the principle of Bayesian techniques based on deep ensembles, but significantly reduces their cost via multiple low-rank perturbations of parameters arising from a pre-trained neural network. Both vanilla version of ensembles as well as more sophisticated schemes such as Bayesian learning with Stein Variational Gradient Descent (SVGD), previously deemed impractical for large models, can be seamlessly implemented within the proposed framework, called \underline{B}ay\underline{e}sian \underline{L}ow-Rank \underline{L}e\underline{A}rning (Bella).
In a nutshell, i)~Bella achieves a dramatic reduction in the number of trainable parameters required to approximate a Bayesian posterior; and ii) it not only maintains, but in some instances, surpasses the performance of conventional Bayesian learning methods and non-Bayesian baselines. Our results with large-scale tasks such as \texttt{ImageNet}, \texttt{CAMELYON17}, \texttt{DomainNet}, \texttt{VQA} with \textsf{CLIP}, \textsf{LLaVA} demonstrate the effectiveness and versatility of Bella in building highly scalable and practical Bayesian deep models for real-world applications.

\textbf{Code:~}\url{https://bnn-bella.github.io/BNN-Bella/}

\end{abstract}


\section{Introduction}
\label{sec:intro}
\begin{figure}
    \centering
    \includegraphics[width=0.99\linewidth]{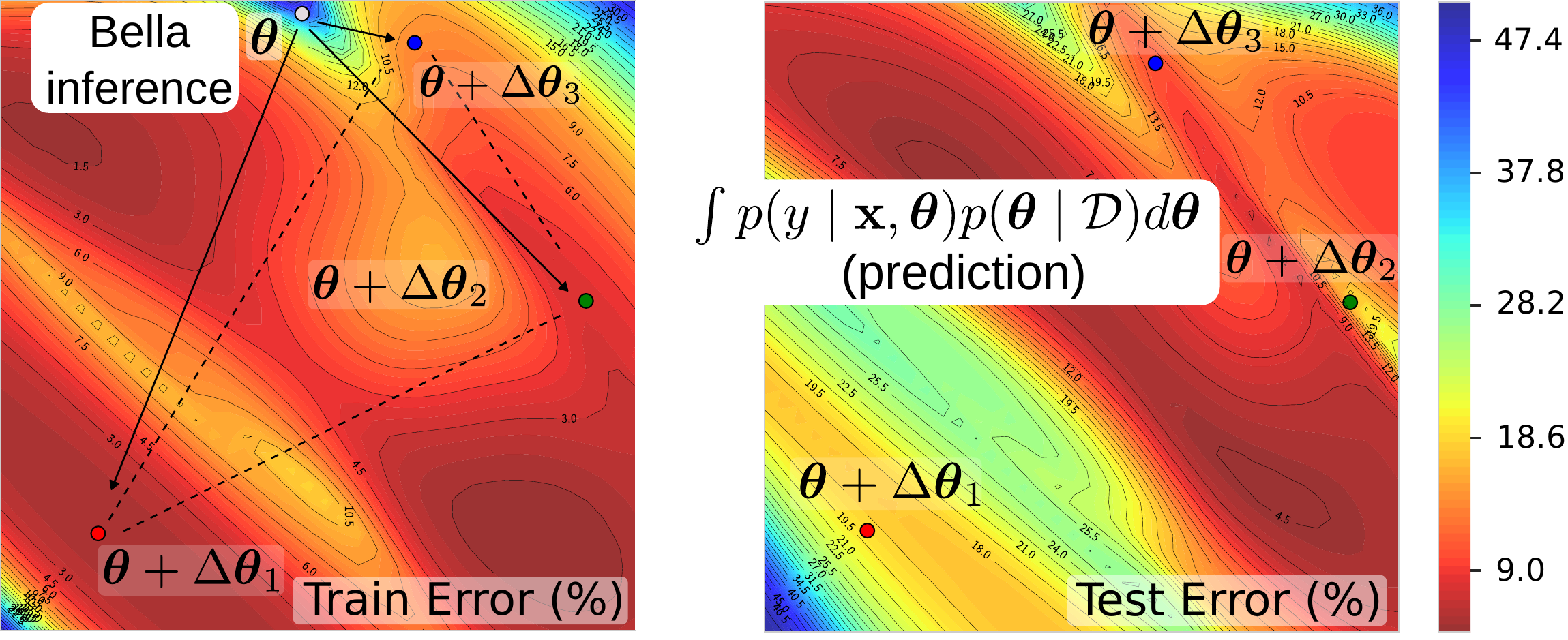}
    \caption{Train and Test Error (\%) landscapes of \texttt{CAMELYON17}. Training landscape demonstrates the learned 3 particle approximation of the modes of the posterior from a pre-trained model $\btheta$---modification of parameters in the constrained region ($\Delta\btheta$) leads to approaching posterior modes in inference. 
    Here, we observe a key benefit of our \method approximation compared to a point estimate---while a single parameter particle (e.g. $\btheta +\Delta\btheta_1$) do not generalize well, Bayesian prediction with \cref{eqn: bma}, effectively an average over multiple parameter settings, leads to better performance.
    }
    \label{fig:error-train-test-cam}
    \vspace{-2mm}
\end{figure}
Bayesian deep learning~\cite{neal2012bayesian} provides mechanisms, for building predictive models more \textit{robust} to adversarial attacks, resilient to unseen or out-of-distribution data and a \textit{theoretical} framework for estimating model uncertainty~\cite{BAL,advbnn,izmailov2020subspace,chen2020projected,wilson2022evaluating,DoanAAIfeature2023,baoesorics2024}. 
Consequently, embracing Bayesian deep learning (BDL) represents a significant stride towards building more reliable and trustworthy AI systems for various real-world applications (e.g., autonomous driving, medical image analysis, etc.). \textit{Unfortunately, their practical use is encumbered by their computational complexity}. 

Unlike traditional alternatives with point estimates---a single set of model parameters mapping inputs to outputs---Bayesian models learn the distribution of model parameters to offer a distribution over possible predictions. 
Consider a neural network $f(\bx,\btheta)$ with input $\mathbf{x}$ parameterized by $\btheta$ and its corresponding prior distribution $p(\btheta)$. The likelihood $p(\D\vert\btheta)$ is determined by $f$ \cite{Bishop2006} and then, Bayesian inference seeks to use Bayes' theorem to derive a {\it posterior} distribution given by
$p\left(\btheta\vert \D\right) = { p\left(\D\vert\btheta\right)p\left(\btheta\right) }/{p(\D)}$
to compute the predictive distribution:
{\small\begin{align}
p(y|\bx,\mathcal{D})=\int p(y\mid\bx,\btheta) p(\btheta\mid\mathcal{D}) d\btheta. \
\label{eqn: bma}
\end{align}}

Despite evidence to support the adoption of Bayesian learning, their widespread use  is hindered by several challenges; \textit{a primary issue is the intractability of the posterior distribution $p(\btheta\vert\D)$ in practical learning tasks}.
The exact solution for the posterior, even for networks of moderate size, is impractical. Because of deep neural network's complexity and the high-dimensional integral of the resulting denominator. 
This necessitates the use of approximations.

Our work expands upon recent research demonstrating the effectiveness of ensemble methods and Stein Variational Gradient Descent (SVGD)~\cite{liu2016stein} approaches in BDL, for example, \cite{deep_ensemble,lakshminarayanan2017simple,abbasnejad2020gold,doan2022bayesian,izmailov2020subspace,chen2020projected}, to \textit{provide a practical method for ensemble Bayesian deep learning for large-scale tasks}. 
To achieve a practical and efficient approach, we consider spawning parameter particles (models) by adding linear interpolation to a pre-trained model's parameters to build an ensemble. To further enhance the efficiency of these interpolations, inspired by the success of low-rank adapters (LORAs)\cite{hu2022lora}, we propose using low-rank parameters in the interpolation process, we refer to as Bayesian Low-Rank Learning (Bella).

To validate this, we compare \method to the full-parameter alternative in terms of (1) accuracy, (2) out-of-distribution generalization, (3) adversarial robustness, and (4) alignment of uncertainty estimation with human confidence. In summary, our findings indicate that this approach achieves comparable performance to full SVGD or ensemble methods at a fraction of the computational cost, with the exception of adversarial robustness. For example, Figure \ref{fig:error-train-test-cam} illustrates the application of our proposed approach of  Bayesian inference in an application with distribution shifts (\texttt{CAMELYON17} \cite{wilds2021} benchmark). The particles (\ie, parameter samples from the posterior) obtained using \method prove effective in improving generalization, improving the common single particle fine-tuning and comparable with full ensemble-based baselines at a fraction of the cost. Our key contributions are: 
\begin{itemize}[leftmargin=3mm]
    \item We propose a new Bayesian learning framework, \method, for SVGD approximation of a posterior---exploiting the availability of pre-trained models, we spawn particles or models by linear interpolation of a constrained set of model parameters for a SVGD approximation of a posterior.
    
    \item Our approach more efficiently captures the complexity and multi-modality of the solution space compared to current SVGD but at a fraction of the cost---we observe on-par performance with full SVGD in uncertainty estimation, performance improvement, and robustness but only use less than 0.3\% of parameters.

    \item We demonstrate \method performs on par or better than baselines---ensembles or current SVGD implementations. \method consistently outperforms the non-BNN counterparts on 8 datasets; 
    image classification (\texttt{ImageNet}, \texttt{CIFAR 10/100}), Out-of-distribution (\texttt{DomainNet}, \texttt{Camleyon17}, \texttt{CIFAR-10-C}), and \texttt{VQA} as well as \textit{adversarial robustness}. Notably, it sets a new state-of-the-art for the \texttt{CAMELYON17} (see Appendix for more details) according to the leaderboard~\cite{leaderboard}.
    
    \item \method employed with the multi-modal model~\textsf{LlaVa}~\cite{liu2023llava}, leads to improved performance and uncertainty estimation highly correlated with human confidence.
\end{itemize}

\textbf{Code:~}\url{https://bnn-bella.github.io/BNN-Bella/}

\section{Related Work}


\heading{Parameter-Efficient Fine Tuning.~}In contrast to fine-tuning all parameters, recent research proposed inserting \textit{adapters} in between existing neural layers to reduce the number of trainable parameters and, subsequently, the compute time (GPU consumption)~\cite{houlsby_parameter-efficient_2019, rebuffi_learning_2017, lin-etal-2020-exploring}.
Hu et al.~\cite{hu2022lora} use a bottleneck structure to impose a low-rank constraint on the weight updates, named LoRA. The key functional difference is that LoRA can be merged with the main weights during inference, thus avoiding the introduction of any latency whilst significantly reducing the number of parameters. 

\vspace{2px}
\heading{Fine-Tuning Approaches and Bayesian Deep Learning.~}
Previous research investigating the application of fine-tuning approaches for BDLs have predominantly focused on large language models (LLMs), e.g. ~\cite{fan2020bayesian,zhang2021bayesian,yang2023bayesian}. Notably, marking a departure from the conventional methods relying primarily on tuning the network's parameters, these studies chose to define priors and approximate posterior over low rank \textit{attention weights}. Concurrently, Yang et al.~\cite{yang2023bayesian} introduced the concept of Laplace LoRA to incorporate Bayesian concepts to enhance the calibration of fine-tuned LLMs. However, Laplace's method~\cite{laplace2021} relies on a Gaussian approximation of the posterior distribution. Whilst this can be effective for unimodal and symmetric distributions, similar to~\cite{dusenberry2020efficient,krueger2017bayesian,vadera2022impact}, the approach does not fully encapsulate the intricacies of more complex posteriors, particularly in neural networks where the posterior has multimodality and asymmetry~\cite{izmailov2021bayesian}. Deep ensembles~\cite{lakshminarayanan2017simple} typically perform better in practice compared with variational and Laplace methods, due to their ability to capture multiple modes. When employing fine-tuning, a direct application of ensembling for LoRAs was considered in \cite{wang2024lora}. Some interpretations of deep ensembles suggest that they approximate gradient flows in function spaces and that building desirable properties into an ensemble (such as repulsive behavior), is possible~\cite{WilGhaSejKno2023}. SVGD~\cite{liu2016stein}, can be viewed in a similar vein. However, while these more sophisticated, repulsive, ensembling approaches are highly impractical in the pre-training phase, we argue that their expressivity can be brought to bear precisely in tandem with low-rank fine-tuning, which is the viewpoint we adopt in this contribution.

Building upon these foundations, our research represents a pioneering effort to apply the principles of repulsive ensemble-based low-rank fine-tuning to computer vision. In particular, we bridge pre-training and fine-tuning phases with recent conjectures on mode connectivity~\cite{git_rebasin}. Our methodology not only capitalizes on the efficiency of fine-tuning techniques, e.g. \cite{opendelta,ding2022delta,dettmers2023qlora} but also innovatively addresses the scalability challenges inherent to BDLs. 
Overall, our work sets a new precedent in applying Bayesian approaches to computer vision tasks by offering a scalable and efficient framework for enhancing model performance.

\section{Stein Variational Gradient Descent Primer}

Bayesian inference techniques have been integral to the development of neural networks, with a rich history underscored by previous works \cite{neal2011mcmc, neal2012bayesian, MacKay1991, welling2011bayesian, blei2017variational}. 
Variational Inference (VI) \cite{blundell2015weight,blei2017variational} and Markov Chain Monte Carlo (MCMC)~\cite{neal2011mcmc,welling2011bayesian} are two primary approximate Bayesian inference frameworks. The former substitutes the true posterior with a tractable alternative while the latter involves sampling. 
However, accurately computing the posterior with either MCMC or VI becomes computationally infeasible when dealing with large-scale networks containing millions of parameters. Although approximations can be obtained more efficiently with VI~\cite{blundell2015weight}, VI is also demonstrably too restrictive to resemble the \textit{multi-modality} of the true posterior and suffers from mode collapse~\cite{izmailov2021bayesian}.

Stein Variational Gradient Descent (SVGD)~\cite{liu2016stein}, among other ensemble-based approaches, is an alternative approximate Bayesian technique that combines the strengths of MCMC and VI by transporting a set of parameter \emph{particles} to fit the true posterior distribution, while encouraging diversity among the particles, by incorporating a repulsive term in the parameter updates. 
This diversity prevents the mode collapse and enables learning multiple models to represent various patterns in the data. 
Using $n$ samples from the posterior (\ie parameter particles), SVGD modifies the gradient descent as: 
{\small \begin{align*}
     \btheta_i  &=   \btheta_i - \epsilon_i \hatff{}^*(\btheta_i) \text{\quad with} \\
\label{eq:svgd}
\hatff{}^*(\btheta) &= \sum_{j=1}^n\big[k(\btheta_j, \btheta)  \nabla_{\btheta_j} \log p(\btheta_j\vert \D)
 -\frac{\gamma}{n}\nabla_{\btheta_j} k(\btheta_j, \btheta)\big]\,. 
\end{align*}}
Here, $\btheta_i$ is the $i$th particle, $k(\cdot, \cdot)$ is a kernel function that measures the similarity between particles and $\gamma$ is a hyper-parameter. 
Notably, the kernel function encourages the particles to be dissimilar in order to capture more diverse samples from the posterior and $\gamma$ controls the trade-off between the diversity of the samples versus the minimization of the loss.


\section{Bayesian Low-Rank Learning (\method) for SVGD}

\label{sec:method}

The problem with the current SVGD and other ensemble-based methods in large deep neural networks is its huge computational cost. This renders it infeasible to train efficiently and to scale to a sufficient number of parameter particles for accurately approximating the posterior distribution, which currently remains coarse.
In this work, we propose to capitalize on the low-rank representations of fine-tuning in order to construct a practical and scalable variant of SVGD. We note that while our approach may not fully capture the diversity of the multi-modal posterior, a recent conjecture~\cite{git_rebasin} suggests that these modes might result from parameter permutations in neural networks, leading to models that are functionally equivalent. Building on this idea, Bella
could serve as a practical alternative with sufficient theoretical justification for ensemble methods. 

Consider any dense layer, for which there is a fixed pre-trained weight matrix $\btheta_0\in \mathbb R^{d_1\times d_2}$ with $d_1, d_2$ the corresponding numbers of hidden units. We consider $n$ low-rank perturbations of $\btheta_0$ as 
{\small\begin{equation}
    \btheta_i = \btheta_0 + \Delta\btheta_i = \btheta_0 + \B_i\A_i\,,\quad i=1,\ldots,n.
\end{equation}}
where $\B_i\in\mathbb{R}^{d_1 \times \nlr}$, $\A_i\in\mathbb{R}^{\nlr \times d_2}$ are the low-dimensional update parameters,  and $r\ll d_1,d_2$ is 
the update's rank.
Now Bella proceeds as the joint SVGD on $(\A_i,\B_i)$, with updates
{\small \begin{align*} \label{eq:bella}
     \A_i  =   \A_i - \epsilon_i \sum_{j=1}^n\hatff{}_j^*(\A_i),~~~
     \B_i  =   \B_i - \epsilon_i \sum_{j=1}^n\hatff{}_j^*(\B_i)     
\end{align*}\begin{align*}
\text{with\quad}\hatff{}_j^*(\B_i) &=  k_{i,j} \! \nabla_{\B_i} p\left(y\mid \bx,\btheta_0 + \B_i\A_i\right) -\frac{\gamma}{n}\nabla_{\B_i} k_{i,j},\nonumber \\
\hatff{}_j^*(\A_i) &=  k_{i,j} \! \nabla_{\A_i} p\left(y\mid \bx,\btheta_0 + \B_i\A_i\right) -\frac{\gamma}{n}\nabla_{\A_i} k_{i,j}\,,
\end{align*}}
where we denote $k_{i,j}=k\left(\B_j\A_j, \B_i\A_i\right)$.
Here, we have placed a zero-mean Gaussian prior on $(\A_i,\B_i)$, but other choices are possible. 
Note that the kernel function on $(\A_i,\B_i)$ is given by $k\left(\btheta_0 + \B_j\A_j, \btheta_0 + \B_i\A_i\right)$, which ensures that the similarity is computed on the original parameter space, and in the commonly used case of shift-invariant kernels, this simplifies to $k\left(\B_j\A_j, \B_i\A_i\right)$. Further simplifications are obtained for specific kernel functions -- in particular, in the case of Gaussian RBF, while the naive implementation would require the cost of $O(rd_1d_2)$ for a single kernel evaluation, we can bring it down to $O(r^2(d_1+d_2))$ using standard trace manipulation, as described in the Appendix. This procedure can be repeated across all dense layers. 

Bella introduces a significant improvement in the efficiency of model training and execution. By utilizing the same pre-trained weights \( \btheta_0 \) across all parameter particles but allowing for individual low-rank adaptations \( \Delta\btheta_i \), we achieve a balance between parameter sharing and the diversity necessary for effective learning. 
Bella significantly reduces the parameter space from the full matrix's \( d_1d_2 \) to just \( \nlr (d_1 + d_2) \), thereby enhancing both efficiency and scalability.
This setup not only reduces the computational burden during training but also streamlines the process at inference time. The heavy lifting is done once by loading the large base model \( \btheta_0 \), and the lightweight low-rank adapters \( \Delta\btheta_i \) can be dynamically applied with minimal overhead in order to approximate the posterior predictive distribution   as  $p ( y^* \mid \mathbf{x}^*, \D ) \approx \frac{1}{n} \sum_{i=1}^{n} p(y^* \mid \mathbf{x}, \boldsymbol{\theta}_{0} + \Delta\btheta_i)$.
 This approach is particularly advantageous in large-scale models, where the weight matrices \( \btheta_0 \) are of substantial dimensions.


\section{Empirical Experiments and Results}
\label{sec:results}


In this section, we provide an in-depth overview of our experimental setup, detailing the methodology, equipment, and procedures employed in the implementation of \method. We aim to compare \method with established ensemble-based methods and SVGD, both of which have demonstrated effectiveness in BDLs~\cite{deep_ensemble,lakshminarayanan2017simple}.

\begin{table*}[b]
\setlength{\tabcolsep}{5pt} 
\centering
\resizebox{1.0\textwidth}{!}{%
\begin{tabular}{@{}l|c>{\columncolor{grey}}ccc|c>{\columncolor{grey}}ccc|>{\columncolor{grey}}c}
\dtoprule
\multirow{2}{*}{\makecell{~~~~~~~~~~~~~~~~~~~~~~~~~Models}}          & \multicolumn{4}{c|}{Bseline Models (SVGD)} & \multicolumn{4}{c|}{\method Models (SVGD)} & \multicolumn{1}{c}{\makecell{Single}}\\ \cline{2-10} 
 &
  \multicolumn{1}{c}{\makecell{$n=3$}} &
  \multicolumn{1}{c}{\makecell{$n=5$}} & 
  \multicolumn{1}{c}{\makecell{$n=20$}} &
  \multicolumn{1}{c|}{\makecell{$n=40$}} &
  \multicolumn{1}{c}{\makecell{$n=3$}} &
  \multicolumn{1}{c}{\makecell{$n=5$}} &
  \multicolumn{1}{c}{\makecell{$n=20$}} &
  \makecell{$n=100$} & \multicolumn{1}{c}{\makecell{$n=1$}}\\ \hline
\multicolumn{1}{c|}{\makecell{ Trainable Parameters}} & 340M      & 567M      & 1.76B & 3.51B &  1.10M    & 1.84M    & 7.37M & 36.86M & 113M\\ \hline
\multicolumn{1}{c|}{\makecell{Memory Consumption (RAM in GB)}}      & 6.71   & 8.35   & 26.08 & 48.45 & 4.48  & 4.50  & 4.63 & 5.19 & 5.05 \\ \hline
\multicolumn{1}{c|}{\makecell{Storage Consumption (MB)}}      & 1321   & 2222   & 8868 & 17735 & 436  & 439  & 460 & 572 & 433 \\ \dbottomrule
\end{tabular}%
}
\caption{Computational cost to train different models based on \textsf{CLIP} architecture for different datasets. Notably, with SVGD Baseline Models, we can only train up to $n$=40 particles on a A6000 48~GB GPU, while we can increase to more than 100 parameter particles with our \method method with negligible increase of GPU consumption. The grey columns correspond to costs (parameters) of models in \cref{tab:results}. 
}
\label{tab:cost}
\end{table*}

\subsection{Experimental Set-up}
\heading{Datasets.~}In this research, we have employed a variety of datasets, each selected for their relevance and contribution, they include \texttt{CIFAR-10}, \texttt{CIFAR-100}~\cite{cifar}, \texttt{CIFAR-10-C}~\cite{hendrycks2019benchmarking}, \texttt{STL-10}~\cite{stl10}, \texttt{CAMELYON17}~\cite{bandi2018detection}, \texttt{ImageNet}~\cite{russakovsky2015imagenet}, and \texttt{DomainNet}~\cite{peng2019moment}. 
We also consider VQA v2 dataset utilized for Visual Question Answering (VQA).
Additional details are in the Appendix.

\begin{table*}[t]
\centering
\resizebox{1.0\linewidth}{!}{%
    \setlength{\tabcolsep}{4pt} 
    \resizebox{\textwidth}{!}{%
    \begin{tabular}{lcc|cc|c|c}
    \dtoprule
    \multirow{2}{*}{Datasets} & \multicolumn{2}{c|}{\method Models} & \multicolumn{4}{c}{Baseline Models (Base)} \\ \cline{2-7} 
                              &  \makecell{Ensemble\\(n=5)}   & \makecell{SVGD\\(n=5)}    & \makecell{Ensemble\\(n=5)}   & \makecell{SVGD\\(n=5)}   &  Single & VBL \\ \hline
    \texttt{CIFAR10}                   &  97.32 $\pm$ 0.37\%      & \textbf{97.57} $\pm$ 0.38\% & 97.26 $\pm$ 0.28\%     & \underline{97.56} $\pm$ 0.31\%  & \textit{96.86} $\pm$ 0.43\% & 94.04 $\pm$ 0.37\% \\ 
    \texttt{CIFAR100}                  &  86.02 $\pm$ 0.48\%      & \textbf{87.63} $\pm$ 0.46\%  & 86.65 $\pm$ 0.24\%     & \underline{87.03} $\pm$ 0.29\%  & \textit{85.14} $\pm$ 1.3\%  & 85.01 $\pm$ 1.21\% \\ 
    \texttt{CAMELYON17}             &93.11 $\pm$ 1.36\%    & \underline{93.61} $\pm$ 1.28\%   & 93.29 $\pm$ 0.94\%     & \textbf{93.98} $\pm$ 1.23\%  & \textit{90.25} $\pm$ 2.31\%     & 91.58 $\pm$ 1.38\% \\ 
    \texttt{DomainNet}             &  80.34 $\pm$ 2.36\%  &  81.41 $\pm$ 3.06\%  &  \underline{82.96} $\pm$ 2.11\%   &  \textbf{83.66} $\pm$ 1.76\% &  \textit{69.75} $\pm$ 4.86\%   & 80.60 $\pm$ 2.37\% \\ 
    \texttt{ImageNet}            &   77.29  &  78.24  & \underline{78.93}    & \textbf{79.36}  & \textit{76.87} & - \\ \dbottomrule
    \end{tabular}
    }%
}

\caption{Comparing \method models with their baseline (\textit{Base}) counterparts on \textit{vision benchmarks}. 
\method models are well-performing and on par with their baseline Ensemble, SVGD, and VBL whilst consuming only a fraction of cost. The best-performing results are in \textbf{bold}, the second-best in \underline{underline}, and the least favorable are in \textit{italic} for emphasis.}
\label{tab:results}
\vspace{-5mm}
\end{table*}

\vspace{2px}
\heading{Neural Architecture.~}In our experiments, we employed the \textsf{CLIP} ViT-B/32 model~\cite{ilharco_gabriel_2021_5143773}, a pre-trained variant utilizing contrastive supervision from image-text pairs, as initially introduced in the seminal \textsf{CLIP} research~\cite{Radford2021LearningTV}. 
We conducted end-to-end fine-tuning of the image encoder, adjusting all model parameters, a strategy typically yielding higher accuracy than training only the final linear layer. 
For both our ensemble-based methods, we consider conventional ensemble and SVGD. We use the average logits (unnormalized outputs)  to produce the output \cite{gontijo2021no}. 
For VQA task, we employ the SoTA LLaVA-1.5-7B~\cite{liu2023llava} to showcase the effectiveness of \method on large-scale network architecture.
Detailed hyper-parameters are in the Appendix.

\subsection{Cost Efficiency}
\label{sec:cost}

Cost comparisons (in terms of trainable parameters, memory and storage) are shown in~\cref{tab:cost}. The performance comparison of our Bella models with their respective base models, as well as with Vision Bayesian Lora (VBL)---a derived Laplace's approximation from~\cite{yang2023bayesian} for vision tasks---is shown in \cref{tab:results}. In particular, the gray columns \cref{tab:cost} indicate the models used to generate the results in~\cref{tab:results}.

Impressively, across a spectrum of benchmark computer vision datasets such as \texttt{CIFAR-10}, \texttt{CIFAR-100}, \texttt{Camelyon17}, and \texttt{ImageNet}, the \method models demonstrate superior performance. This is achieved with only a fraction of cost (trainable parameters) as shown in~\cref{tab:cost}, underscoring the models' proficiency in parameter efficiency without compromising on accuracy. Further, the \method models surpass the performance of \textit{Single} models, while employing a comparable level of computational resources---see the Memory Consumption of models used, in \cref{tab:cost} reporting approximately 4.5~GB for \method compared to 5.05~GB for Single. The results across the benchmarks attest to the efficacy of our methodology.

Notably, in \cref{tab:results} with \method SVGD, with 1.6\% of the trainable parameters in comparison to the Single baseline, leads to approximately 2\% and 10\% increase in performance on the OOD tasks of \texttt{CAMELYON17} and \texttt{DomainNet}, respectively; whilst achieving comparable performance with the current SVGD implementation (SVGD baseline model).

Along with \cref{tab:results}, \cref{tab:cost} demonstrates the primary benefit of our approach---\textit{the significant reduction in memory and storage needs}. For example, as seen in \cref{tab:cost},
we obtain approximately a 5$\times$ reduction in model size---that is from 2,222~MB for a 5 particle SVGD baseline model to just 439~MB for a \method SVGD model with 5 particles. This efficiency not only reduces the demands on GPUs but also minimizes potential I/O bottlenecks.

Moreover, the reduced GPU demand, as shown in~\cref{tab:cost}, facilitates larger mini-batch sizes during training to speedup the training process. More crucially, it enables one to enhance the Bayesian posterior's parameter particles to over 100, a feat previously unattainable with current SVGD implementations. 
Significantly, constructing a 100 parameter Bayesian approximation consumes only 5.19~GB memory compared to current SVGD implementations for Bayesian models (SVGD base) needing over 6~GB for \textit{even a mere 3 particle approximation}.

\rqq{Interestingly, our results for \method models are on-par with SVGD and ensemble approximations of the posterior. This provides empirical evidence that with constrained model parameters, it is still possible to reach the diverse modes of the posterior. Further, \textit{the results support recent conjectures on mode connectivity}~\cite{gueta2023knowledge}.}

\begin{table*}[b]
\centering
\setlength{\tabcolsep}{10pt} 
\resizebox{0.96\textwidth}{!}{%
\begin{tabular}{lcccccccccc}
\dtoprule
\multirow{2}{*}{Models} & \multicolumn{5}{c}{\texttt{DomainNet}} & \multirow{2}{*}{Avg.} & \multicolumn{3}{c}{\texttt{CIFAR-10-C}} & \multirow{2}{*}{Avg.} \\ \cline{2-6} \cline{8-10} & Real  & Clip-Art & Infograph & Paint & Sketch &  & Gaussian Blur & Pixelate & Spatter & \\ \midrule
Single base            & 74.61 & 55.29    & 31.81     & 53.87 & 43.84  & 51.89 & 90.58        & 77.94    & 92.14   & 78.53 \\ \hline
VBL            & 82.97 & 59.94    & 26.56     & 52.81 & 46.96  & 53.85 & 89.34        & 76.43    & 89.21   & 70.76 \\ \hline
Ensemble \method         & 82.70 & 61.22    & 28.15     & 55.32 & 51.58  & 55.60 & 93.30        & 85.86    & 94.83   & 82.24 \\ 
Ensemble base          & 85.07 & 65.40    & 36.07     & 57.90 & 54.22  & 59.73 & 91.92        & 86.49    & 93.45   & 84.61 \\ \hline
SVGD \method             & 84.47 & 63.67    & 32.22     & 56.83 & 54.86  & 58.41 & 94.05        & 88.70    & 94.85   & 83.77 \\ 
SVGD base              & 85.42 & 65.53   & 36.58     & 58.18 & 55.47 & 60.24 & 93.02        & 86.75    & 95.19   & 86.84 \\ \dbottomrule
\end{tabular}
}
\caption{{The out-of-distribution generalization performance of \method, measured by accuracy (for number of particles $n=5$). Each column represents a specific shift, either real (\texttt{DomainNet}) or artificial (\texttt{CIFAR-10-C}).}} \label{tab:domainnet_cifar10-c}
\end{table*}

\subsection{Out-Of-Disrtibution (OOD) Datasets}
\label{sec:ood}

Assessing the robustness of machine learning systems to unseen conditions is crucial, particularly regarding their ability to generalize to out-of-distribution (OOD) data. We evaluate robustness to OOD by utilizing multiple OOD benchmarks and estimating accuracy under various noise levels (distribution shifts) to further assess the performance of different baselines in these challenging scenarios.

First, we use the \texttt{DomainNet} dataset, which is one of the most diverse domain adaptation datasets and spans a wide range of visual styles, from real images to abstract art. This variety provides a challenging test bed for algorithms aiming to bridge different visual domains. Our study involves training the \textsf{CLIP} network on the `Real' subset of \texttt{DomainNet} and evaluating its generalization across various domains. Additionally, we use \texttt{CIFAR-10-C}, a corrupted version of \texttt{CIFAR-10}, to further assess the generalization and robustness of models trained on \texttt{CIFAR-10} datasets. The results for both datasets are presented in \cref{tab:domainnet_cifar10-c}.

Furthermore, the \texttt{STL-10} dataset, which has significant label overlap with \texttt{CIFAR-10}, serves as a relevant OOD test case for \texttt{CIFAR-10} models (see the Appendix).

Results from \cref{tab:domainnet_cifar10-c} and the Appendix show that \method models, demonstrating significantly better efficiency, achieve competitive performance compared to more resource-intensive implementations with Ensemble and SVGD baselines across both \texttt{DomainNet} and \texttt{CIFAR-10-C} benchmarks. All Bayesian approximations, including our scalable and efficient method, outperform the single model baseline.

\subsection{Comparing Uncertainty Estimations}

Bayesian models capable of providing a theoretical basis for measuring  \textit{model uncertainty}. Also known as epistemic uncertainty, refers to uncertainty stemming from limitations in our knowledge or understanding of the underlying data generating process or the model itself. One of the ways to quantify model uncertainty is through mutual information estimates, following ~\cite{gal2016uncertainty}.

\heading{Mutual Information (MI).~}
This is the mutual information between the output prediction and the posterior over model parameters $\btheta$, and can be used as a measure of epistemic (model) uncertainty.
It can be expressed as:
$\text{MI}(\btheta, y \mid \mathcal{D}, \bx) = H[p(y\mid \mathcal{D}, \bx)] - \mathbb{E}_{p(\btheta \mid \mathcal{D})} H[p(y \mid  \btheta,\bx)]
$
If the parameters at input are well defined (\,e.g., data seen during training), then we would gain little information from the obtaining label, or the MI measured will be low. 

We employ MI to measure uncertainty to investigate whether the \method approximations of the posterior leads to uncertainty estimates commensurate with those obtained from SVGD baselines. This provides empirical evidence of a functional equivalence of the \method approximations of the posterior to that obtained from the current computationally intensive implementation of SVGD.

\heading{Datasets.~}We utilize the \texttt{CIFAR-10-C} task, featuring corrupted images, to examine the uncertainty of model predictions trained on the standard \texttt{CIFAR-10} dataset. Additionally, we assess the uncertainty measures on the \texttt{CAMELYON17} dataset, which is characterized by inherent dataset shifts within itself.

\begin{figure*}[t]
    \centering
    \includegraphics[width=\linewidth]{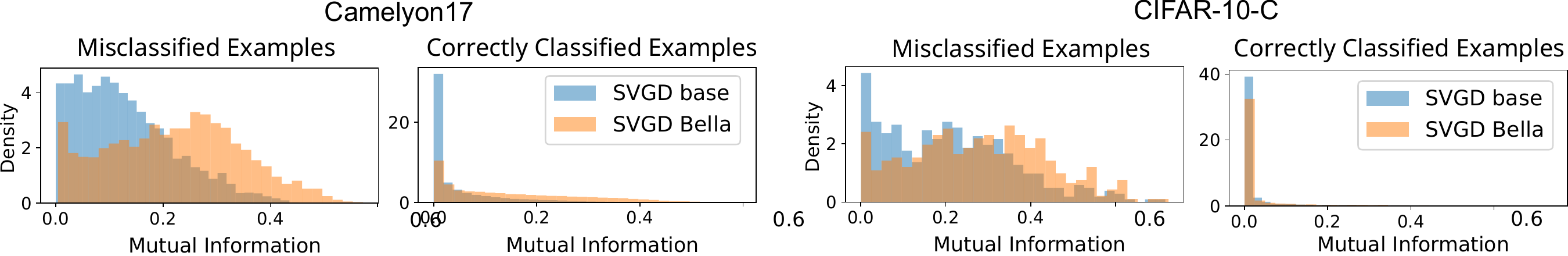}
    \caption{{Evaluation of uncertainty estimations using Mutual Information on \texttt{CAMELYON17} and \texttt{CIFAR-10-C} datasets. $\uparrow$ MI for Misclassified Examples is better---denoted by the distribution shifting $\rightarrow$. In contrast,  $\downarrow$ MI for Correctly Classified Examples is better---denoted by the distribution shifting $\leftarrow$. 
    }
    }
    \label{fig:mi-cifar-c}
    \vspace{-2mm}
\end{figure*}

\heading{Results.~}~\cref{fig:mi-cifar-c} demonstrates the effectiveness of our approach to estimate uncertainty. Our \method perform similarly to the SVGD base model, with a slightly better uncertainty on misclassified images of \texttt{CAMELYON17} and corrupted \texttt{CIFAR-10-C} datasets (under brightness corruption with the maximum intensity), see details in the Appendix.
\vspace{-2mm}


\subsection{Robustness against Adversarial Examples}
\label{sec:robustness}

\begin{figure}[h]
    \centering
    \includegraphics[width=.7\linewidth]{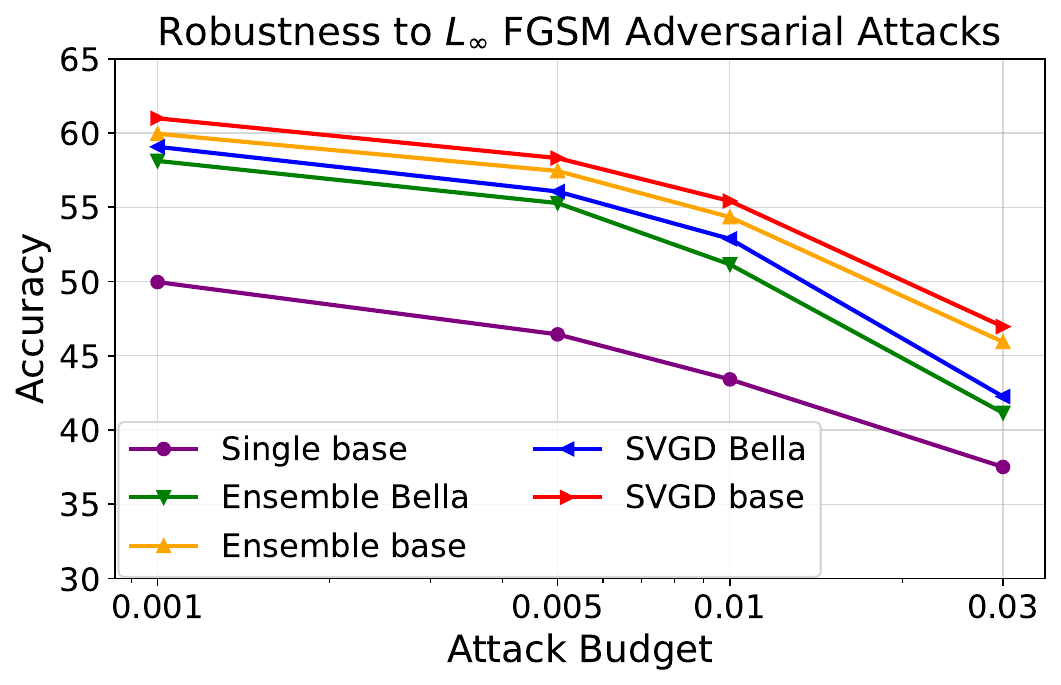}
    \caption{Comparison of model robustness to $L_{\infty}$ FGSM adversarial attacks across varied attack budgets on the \texttt{CIFAR-10} dataset. 
    }
    \label{fig:robustness}
    \vspace{-3mm}
\end{figure}

In this section, we examine the resilience of our proposed \method against adversarial attacks, specifically employing the $L_{\infty}$ Fast Gradient Sign Method (FGSM) across various attack budgets as detailed in~\cref{fig:robustness}. This analysis aims to benchmark the robustness of our method in comparison to traditional models under adversarial conditions. We employ the robustness benchmark~\cite{cleverhans} to deploy the attack on \texttt{CIFAR-10} test set and report results in~\cref{fig:robustness}.

The findings presented in~\cref{fig:robustness} reveal that conventional models such as SVGD and Ensemble exhibit just slightly greater resistance to adversarial attacks. We attribute this enhanced robustness to the broader diversity in model parameters, which stems from their capacity to adjust the entire network's parameters, unlike the \method models.

Significantly, despite operating within the same computational constraints as a singular network model, our \method demonstrates enhanced efficacy in mitigating adversarial attacks, thereby bolstering its robustness.

\subsection{Ablation Studies}
\label{sec:ablation}

This section undertakes a series of ablation studies to examine the effects of various components within \method. 
Our analysis includes an exploration of the training costs associated with different ranks and their consequent influence on model performance. Given that \method incorporates multiple parameter particles, we also delve into how varying the number of these particles affects \method's efficacy. 
Additionally, we explore the application of low-rank adapters across different layers and assess their impact. Further details on other studies are in the Appendix. We show in the Appendix that we achieve state-of-the-art performance on \texttt{CAMELYON17}.

\heading{Ablations on rank $\nlr$.~}As outlined in~\cref{sec:method}, we substitute the network's extensive full matrix with low-rank matrices defined by the `rank' parameter ($\nlr$). This section aims to assess how this parameter influences \method's performance.

\begin{figure}[t]
  \centering
  \begin{minipage}{0.5\columnwidth}
    \centering
    \includegraphics[width=\textwidth]{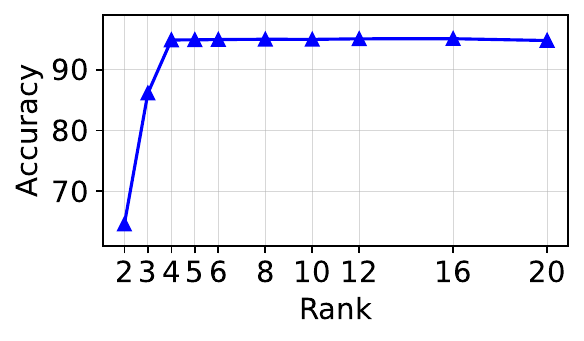}
  \end{minipage}%
  \hfill
  \begin{minipage}{0.5\columnwidth}
    \centering
    \includegraphics[width=\textwidth]{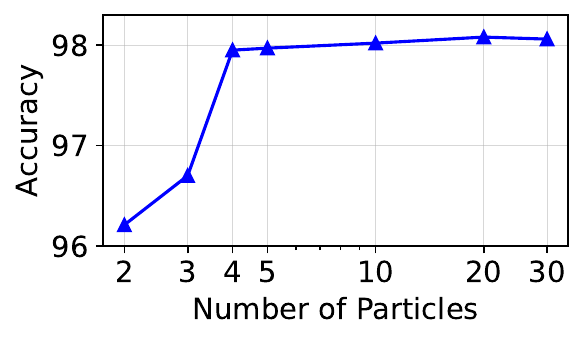}
  \end{minipage}
  \vspace{-1mm}
  \caption{The impact of ranks on \texttt{CAMELYON17} performance (left), as well as the impact of the number of parameter particles on \texttt{CIFAR-10} (right) on \method performance.}
  \label{fig:ablation_rank}
  \vspace{-3mm}
\end{figure}

To assess the influence of rank, we conduct an analysis on the challenging, large-scale, \texttt{CAMELYON17} task. The results, reported in~\cref{fig:ablation_rank} (left), elucidate the relationship between rank size and performance. While utilizing a smaller rank considerably reduces the parameter space, it also restricts \method's learning capacity, hindering its ability to achieve optimal performance. In contrast, increasing the rank to 4 significantly enhances performance. However, performance tends to plateau at a rank of 16, indicating a saturation point.

\heading{Ablations on the number of particles $n$.~}In this section, we delve into the influence of the quantity of parameter particles on the performance of \method. The findings, depicted in the right in~\cref{fig:ablation_rank}, reveal an improvement in \method's performance with an increase in the number of particles. 

This outcome is both intuitive and insightful, as a larger ensemble of parameter particles enhances the approximation of the Bayesian posterior more effectively.


\vspace{2px}
\heading{Ablations on the number of trainable parameters.~}
We will compare the number of trainable parameters between \method and full SVGD in model performance. Critically, higher number of trainable parameters means higher cost to train. To show generalization, we also employ another large-scale challenging dataset Imagenet in this experiment. 

\begin{figure}[t]
    \centering
    \includegraphics[width=.65\linewidth]{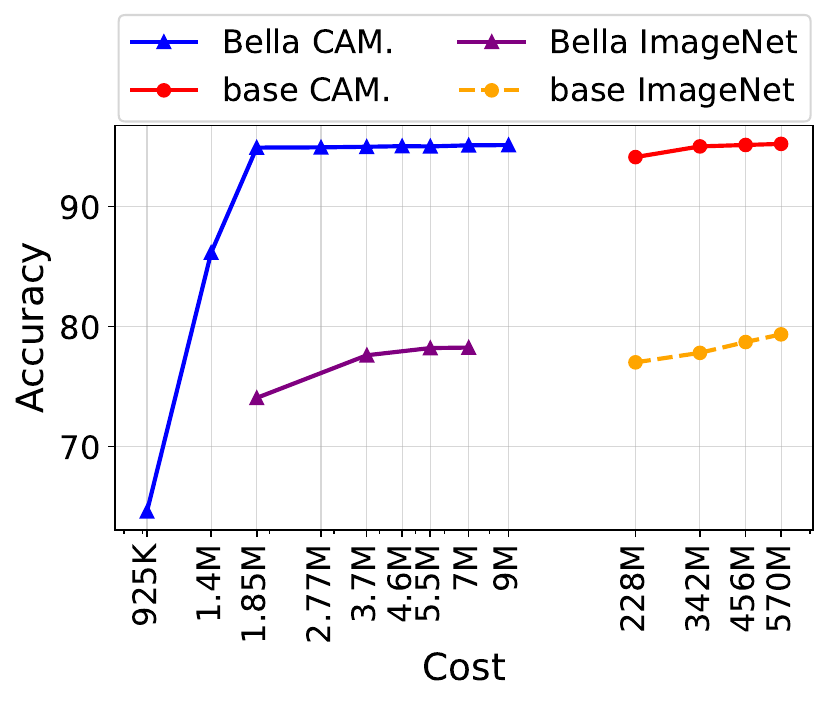}
    \caption{\method achieves similar Accuracy with full SVGD (Base) only with a fraction of cost.}
    \label{fig:Bella-cost}
    \vspace{-3mm}
\end{figure}

\begin{table}[!b]
\vspace{-2mm}
\centering
\setlength{\tabcolsep}{3pt} 
\renewcommand{\arraystretch}{1.1} 
\small 
\resizebox{0.8\linewidth}{!}{%
\begin{tabular}{c|l|ll|lllll}
\toprule
\multirow{2}{*}{\textbf{Data.}}   
  & \multirow{2}{*}{\textbf{Metric}} 
  & \multicolumn{2}{c|}{\textbf{Bella}} 
  & \multicolumn{5}{c}{\textbf{Base}} 
  \\ \cline{3-9} 
  & 
  & Ens. & SVGD 
  & Single & VI & SGLD & Ens. & SVGD 
  \\ 
\midrule

\multirow{4}{*}{\rotatebox[origin=c]{90}{\texttt{CIFAR10}}}
  & ECE  $\downarrow$   & 1.4     & 1       & 8.1   & 0.99 & \underline{0.94} & 2.5   & \textbf{0.55}  
  \\ 
  & MCE  $\downarrow$   & 65      & 26      & 35    & 33   & 24              & \underline{23}   & \textbf{21}    
  \\ 
  & Brier $\downarrow$  & 0.40    & \textbf{0.32}  & 0.49  & 0.43 & 0.36  & 0.49  & \underline{0.33}  
  \\ 
  & AUROC $\uparrow$    & \underline{99.98} & \textbf{99.99} & 99.78 & 99.93 & 99.94 & \textbf{99.99} & \textbf{99.99}
  \\ 
\midrule

\multirow{4}{*}{\rotatebox[origin=c]{90}{\texttt{CIFAR100}}}
  & ECE $\downarrow$    & 5.2     & 4.9     & 6.5   & 4.2  & \underline{3.8} & 5.1   & \textbf{3.3}   
  \\ 
  & MCE $\downarrow$    & \underline{19} & \textbf{15} & 86    & 47   & 41             & 48    & 46    
  \\ 
  & Brier $\downarrow$  & 0.23    & \underline{0.19} & 0.23  & 0.2  & \underline{0.19} & \underline{0.19} & \textbf{0.18}  
  \\ 
  & AUROC $\uparrow$    & \textbf{99.88} & \textbf{99.88} & 99.71 & \underline{99.87} & 99.83 & \textbf{99.88} & \textbf{99.88}
  \\ 
\midrule

\multirow{4}{*}{\rotatebox[origin=c]{90}{\texttt{CAMELYON}}}
  & ECE $\downarrow$    & 5.5     & 5.3     & 6.4   & \underline{5} & 5.2   & 5.3   & \textbf{4.9}   
  \\ 
  & MCE $\downarrow$    & 31      & \underline{29} & 34    & 36   & 31    & 30    & \textbf{21}    
  \\ 
  & Brier $\downarrow$  & 5.2     & \underline{5.1} & 5.5   & 5.4  & 5.4   & \underline{5.1} & \textbf{4.5}   
  \\ 
  & AUROC $\uparrow$    & 98.4    & \underline{98.7} & 98.12 & 98.6 & 98.6  & 98.5  & \textbf{98.81} 
  \\ 
\midrule

\multirow{4}{*}{\rotatebox[origin=c]{90}{\texttt{\makecell{Domain\\Net}}}}  
  & ECE $\downarrow$    & 5.9     & 5.8     & 7.9   & 5.1  & \underline{4.6} & 5.2   & \textbf{4.9}   
  \\ 
  & MCE $\downarrow$    & 40      & \underline{37} & 61    & 40   & 41    & \underline{37} & \textbf{34}    
  \\ 
  & Brier $\downarrow$  & 5.3     & 5.2     & 6.3   & \underline{5} & 4.4   & \underline{5}  & \textbf{4.7}   
  \\ 
  & AUROC $\uparrow$    & 98.31   & 98.46   & 98.1  & 98.99 & \textbf{99.21} & 98.88 & \underline{99.01}
  \\ 
\bottomrule                            
\end{tabular}}
\caption{Calibration comparison of Bella models (SVGD and Ensemble) with baseline methods across multiple datasets. 
}
\label{tab:perform}
\renewcommand{\arraystretch}{1}
\end{table}

The plot in~\cref{fig:Bella-cost} for the \texttt{CAMELYON17} dataset begins with \method at $\nlr=2$ (comprising $5$ particles) and concludes with full SVGD (SVGD base) using the same number of particles. As $\nlr$ increases, the accuracy of Bellas improves, eventually plateauing at approximately 95.08\%, which is comparable to the accuracy of full SVGD (95.21\%). This highlights the efficiency and advantages of our proposed \method. Remarkably, \method achieves performance on par with full SVGD models despite utilizing a significantly smaller pool of trainable parameters—approximately 0.3\% for $\nlr=4$—compared to the more parameter-intensive alternatives.

\vspace{2px}
\heading{Calibration Study.} Table~\ref{tab:perform} presents a comparison of the Bella models (SVGD and ensemble) against baseline models, including Variational Inference (VI)~\cite{kim2023bayesdll} and Stochastic-Gradient Langevin Dynamic (SGLD)~\cite{welling2011bayesian}, using several key metrics: Expected Calibration Error (ECE), Maximum Calibration Error (MCE), Brier score, and Area Under the Receiver Operating Characteristic curve (AUROC). Lower ECE and MCE values indicate that the model's probability estimates are more reliable. A lower Brier score reflects better accuracy in the model's probability predictions, while a higher AUROC demonstrates superior discrimination between classes. Across the datasets, the Bella models demonstrate performance comparable to the baseline models. This suggests that the Bella models, while utilizing far fewer parameters for training, offer a strong alternative to traditional baseline approaches while maintaining competitive calibration and Discriminative capability.

\subsection{Generalization to a Visual Question Answer}
\label{sec:llava}
In this section, we extend the application of our \method to another challenging vision task, Visual Question Answering (VQA), as detailed in~\cite{vqa_v1}. We leverage the state-of-the-art, pre-trained, large multi-modal model LlaVA ~\cite{liu2023llava,liu2023improvedllava} for this purpose. Utilizing LlaVA transforms VQA into a process where an image and a natural-language question are inputs, and the model generates a free-form, open-ended text answer. Answering questions in VQA requires various intelligence capabilities, including not only image recognition but also complex reasoning. For this task, we employ
\texttt{VQA v2}~\cite{vqa_v1} dataset containing 204,721 images, more than 1 Billion (1B) questions and 10B ground-truth answers in total. There are three main types of answers: Yes/No, a Number, and Other. 

\vspace{2px}
\noindent\textbf{Model.~}We employed our proposed \method on top of LlaVA-1.5-7B~\cite{liu2023llava,liu2023improvedllava}. Further details about the dataset, model and metrics are deferred to the Appendix.

\vspace{2px}
\noindent\textbf{Accuracy.~}
The primary outcomes for Yes/No and Number answer queries are detailed in \cref{tab:vqa-llava}. We chose these question types to ensure a fair comparison and avoid semantic mismatches in open-ended answers in Others.
A distinctive feature of \method is its reduced uncertainty estimates for correct predictions, coupled with increased uncertainty estimates for incorrect ones. Moreover, it surpasses the Single base model regarding Accuracy and Exact Match metrics.

Notably, our method is efficient, particularly in contrast to the resource-intesive baseline Bayesian models tailored for this task (\eg \cite{gade,gade2,abbasnejad2020gold}). The computational requirements for using these baselines render the application of full SVGD or ensemble alternatives impractical, thereby highlighting the impact on practical applications by harnessing the effectiveness of \method. 

        
        
        
        

\begin{table}[t]
    \centering
    \resizebox{0.9\linewidth}{!}{%
        \begin{tabular}{lcccc}
            \multicolumn{5}{c}{VQA: Evaluation of Yes/No Questions} \\
            \toprule
            Models              & Ent. Corrects ($\downarrow$) & Ent. Incorrect ($\uparrow$)     
                                & Acc. ($\uparrow$) & Match ($\uparrow$)\\ 
            \midrule
            Single base         & 0.3336 & 0.5920 & 91.59 & 86.74 \\
            Ens. \method    & 0.3438 & 0.5935 & 91.20 & 86.21 \\
            SVGD \method        & \textbf{0.3245} & \textbf{0.5950} 
                                & \textbf{92.46}  & \textbf{87.83} \\
            \bottomrule
            \multicolumn{5}{c}{VQA: Evaluation of Number Questions} \\
            \toprule
            Single base         & 0.4148 & 0.9911 & 58.69 & 49.26 \\
            Ens. \method    & 0.4248 & \textbf{0.9929} & 57.86 & 48.68 \\
            SVGD \method        & \textbf{0.4059} & 0.9816 & \textbf{60.19} & \textbf{50.99}\\
            \bottomrule
        \end{tabular}%
    }
    \caption{LLaVA-VQA Results: evaluating the accuracy and entropy of correct vs.\ incorrect predictions (Ens: Ensemble).}
    \label{tab:vqa-llava}
\end{table}

\vspace{2px}
\heading{Model Uncertainty and Human Confidence.~}Further, we investigate the relationship between model uncertainty attained from \method and human confidence, facilitated by multiple annotators in the VQA dataset.  For this, we measure the model disagreement by measuring the correlation
between predictive entropy and human annotations of the answers.


\begin{figure}
    \centering
    \includegraphics[width=1.0\columnwidth]{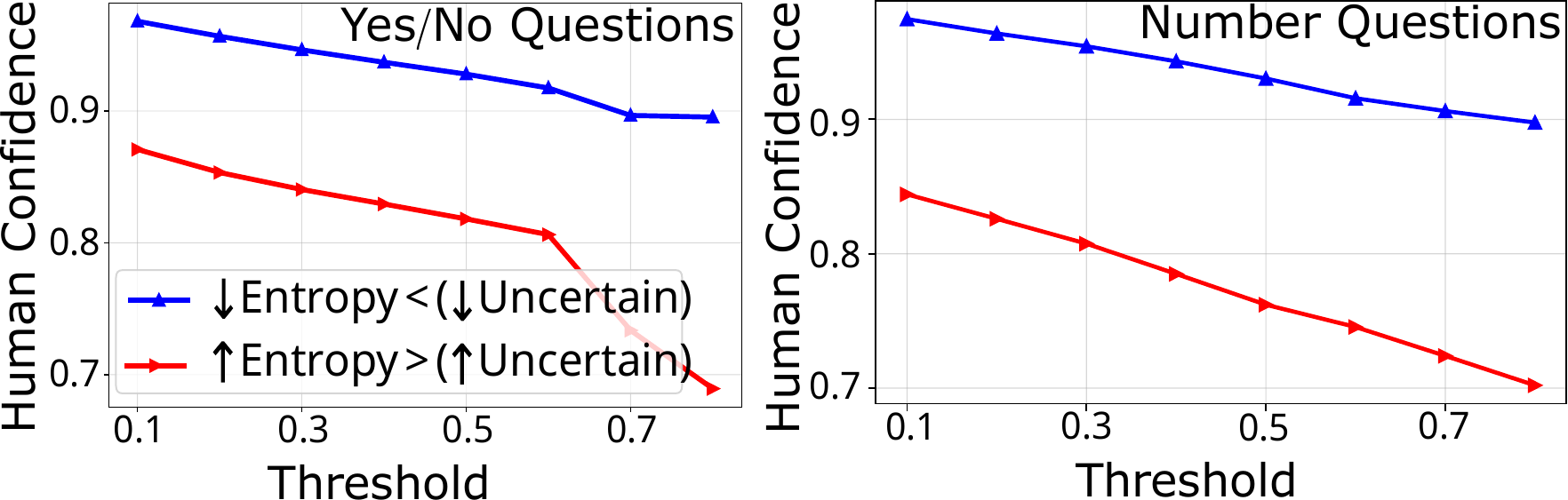}
    \caption{Correlation between model certainty and human confidence. This negative correlation suggests that the model's entropy can serve as a reliable gauge of certainty, mirroring human judgment.
    }
    \label{fig:correlation}
    \vspace{-5mm}
\end{figure}

In \cref{fig:correlation} we show the correlation between the entropy of the model outputs and human confidence levels. By setting a specific threshold for the model's entropy, we effectively bifurcate our predictions into two distinct categories: those with lower entropy fall into the `Low Entropy' group, signaling reduced uncertainty within the model's assessments, while predictions exceeding the threshold are allocated to the `High Entropy' group, indicative of greater uncertainty. Intriguingly, our observations reveal a consistent negative correlation between the model's entropy levels and human confidence across the different query types (Yes/No and Number questions). This pattern suggests that the model's entropy can serve as a reliable gauge of certainty, mirroring human judgment in its response to varying levels of uncertainty.


\section{Conclusion}



We present Bella, an innovative efficient Bayesian Neural Network (BNN) approximation using a base pre-trained model. Bella achieves remarkable compatibility with full-rank BNN methods like SVGD and Ensembles, surpassing single-network solutions in classification, OOD generalization, and uncertainty quantification. This approach enables scalable, reliable BNN implementations, demonstrating the benefits of simple and efficient training over traditional single fine-tuning techniques. Code released at \url{https://bnn-bella.github.io/BNN-Bella/}


\bibliographystyle{unsrt}

\newpage
\appendix

\clearpage
\centerline{{\Large \textbf{\underline{B}ay\underline{e}sian \underline{L}ow-Rank \underline{L}e\underline{A}rning (Bella):}}}
\centerline{{\Large \textbf{A Practical Approach to Bayesian Neural Networks}}}

\section*{Overview of Materials in the Appendices}


\noindent{We provide a brief overview of the extensive set of additional experimental results and findings in the Appendices that follows.} 
\vspace{2mm}

\begin{enumerate}
    \setlength{\itemsep}{3px}
    \item Comparing diversity measures between SVGD \method and its SVGD base model (\cref{appd:diversity})
    \item The generalization of CIFAR-10 pre-trained models on OOD dataset (STL-10) (\cref{appd:ood})
    \item Uncertainty measures on the corrupted CIFAR-10-C dataset between SVGD Bella models and  SVGD models (\cref{appd:mi_cifar10-c})
   
     \item Generalization Performance of Models on \texttt{CIFAR-10-C} Task (\cref{appd:cifar10c})
           \item Additional comparison studies. Including \textbf{SoTA results on~\cite{leaderboard}} (\cref{appd:additional})
       
    \item Impact of $\gamma$ parameter values on robustness of models (\cref{sec:rob-alphas})
   
    \item Detailed information regarding RBF kernel computation in SVGD (\cref{appd:rbf})

     \item Detailed information about VQA task (\cref{appd:llava})
    \item Detailed information regarding datasets and hyper-parameters utilized in the paper (\cref{sec:parameters})
\end{enumerate}

\section{Diversity Measures}
\label{appd:diversity}

This section assesses the diversity of the trained \method, comparing it to the foundational SVGD model. The goal is to determine if \method can maintain the diversity level of the SVGD base model while utilizing significantly fewer parameters.

Given the lack of a conventional metric for evaluating the diversity among parameter particles, we suggest employing the Kullback–Leibler (KL) Divergence. This measure compares the expected parameters of a Bayesian model to the softmax output for each parameter particle, serving as an indicator of model diversity. We calculate this divergence across 10,000 test samples from the CIFAR-10 dataset. In particular,
$$\text{Diversity}=\frac{1}{N}\sum_{i=1}^N{KL\Big[\mathbb{E}_{\btheta}[p(y\mid\bx_i,\btheta)],p(y\mid\bx_i,\btheta)\Big]}$$
where $\text{KL}$ is the Kullback–Leibler divergence, N is the number of samples. 

\begin{figure}[h]
    \centering
    \includegraphics[width=.7\linewidth]{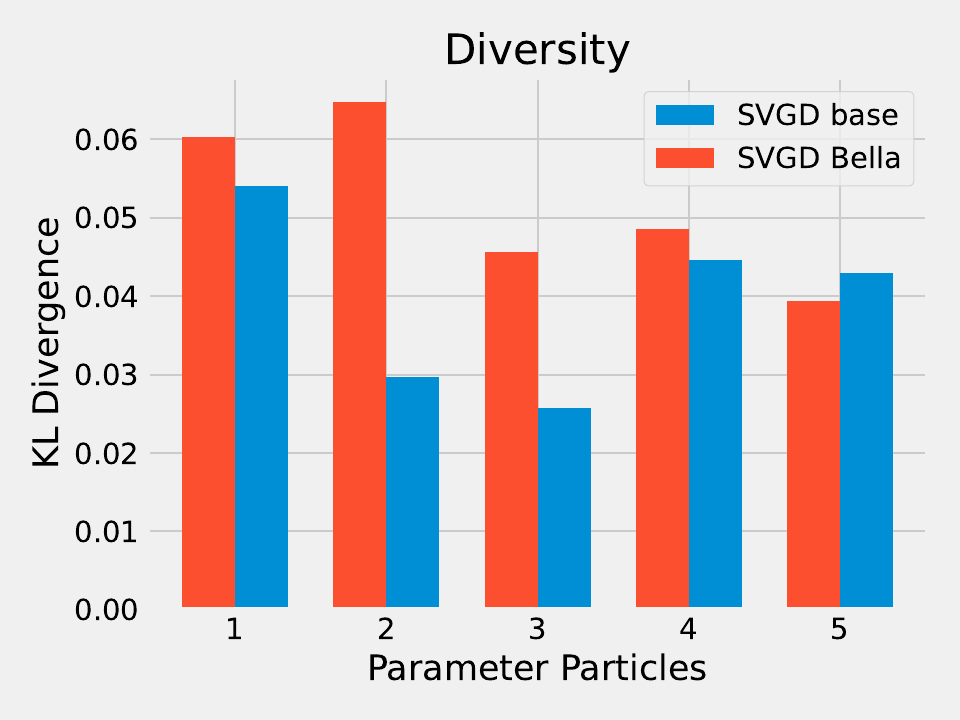}
    \caption{\small{Diversity comparison between SVGD base and SVGD \method. SVGD \method show comparable and slightly better diversity measure compared to its SVGD base model.}}
    \label{fig:diversity}
\end{figure}

\vspace{2mm}
\heading{Results.~}\cref{fig:diversity} presents intriguing findings, where the SVGD \method exhibits comparable and even greater diversity compared to the base SVGD model. We speculate that this increased diversity may stem from the \method's use of significantly fewer trainable parameters, reducing the likelihood of overfitting and facilitating the identification of parameter particles with enhanced diversity, all while preserving model performance. These results further underscore the efficiency and effectiveness of our method in implementing a Bayesian neural network with substantial diversity.


\section{Generalization of CIFAR-10 Models on OOD dataset (STL-10)}
\label{appd:ood}

In this section, we evaluate the generalization of networks trained on \texttt{CIFAR-10} on a \textit{similar} datasets, \texttt{STL-10}, as in~\cite{miller2021accuracy}.

\vspace{2mm}
\heading{STL-10 dataset.~}The \texttt{STL-10} dataset is a benchmark for image recognition algorithms, containing 5,000 labeled training images, 8,000 labeled test images across 10 classes, and 100,000 unlabeled images for unsupervised learning. Images are 96x96 pixels, larger than those in similar datasets like \texttt{CIFAR-10}, facilitating more detailed models and promoting unsupervised learning techniques.

\vspace{2mm}
\heading{Experimental Set-up.~}We choose eight labels in STL-10 test set, same labels observed in \texttt{CIFAR-10} dataset to set a OOD dataset. This includes 6,400 images acting as the OOD dataset for \texttt{CIFAR-10} pre-trained models.

\begin{figure}[!h]
    \centering
    \includegraphics[width=\linewidth]{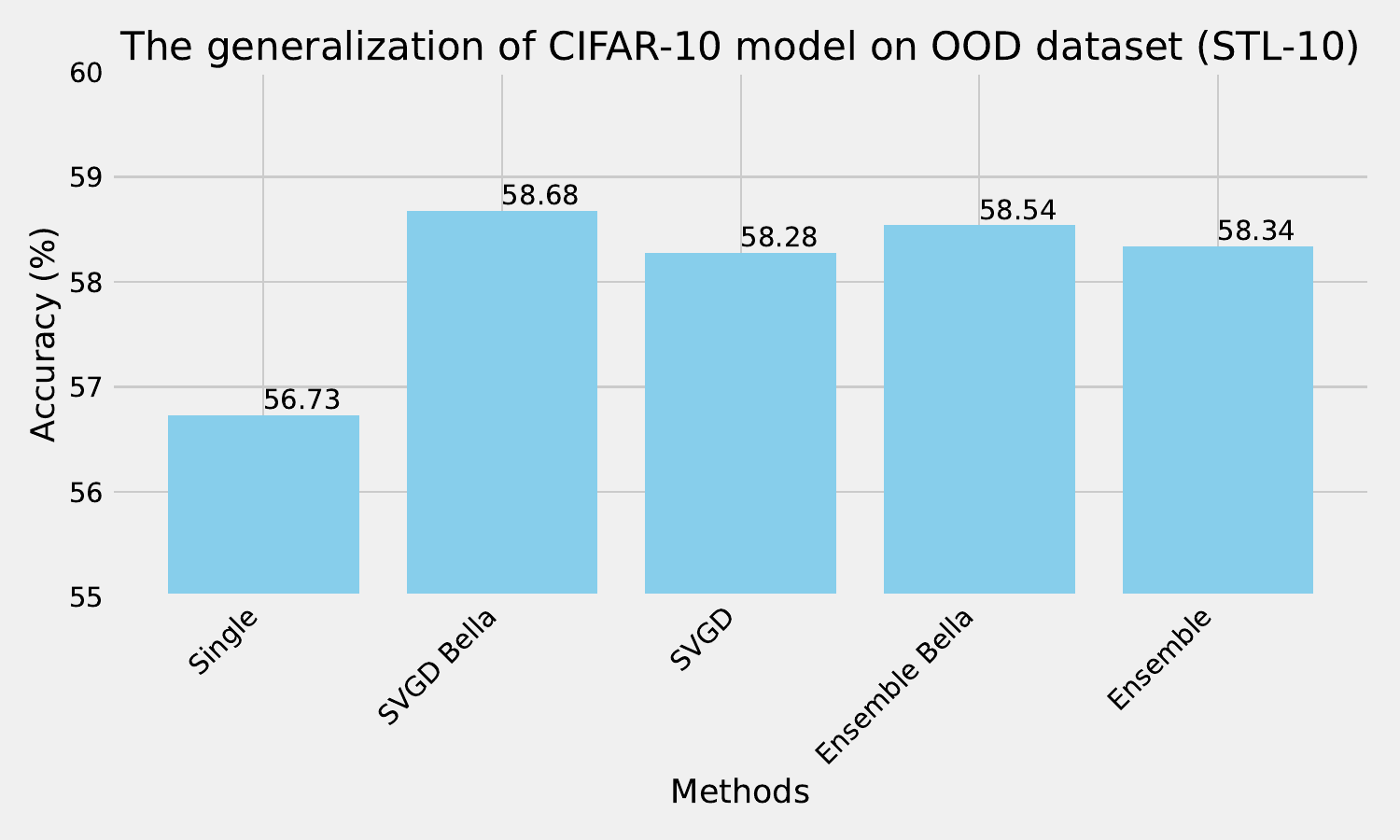}
    \caption{\small{Enhanced OOD detection: \method models show better performance than their counterparts on STL-10 dataset with shared \texttt{CIFAR-10} labels.}}
    \label{fig:ood-stl}
\end{figure}

\begin{figure*}[!h]
    \centering
    \includegraphics[width=0.95\textwidth]{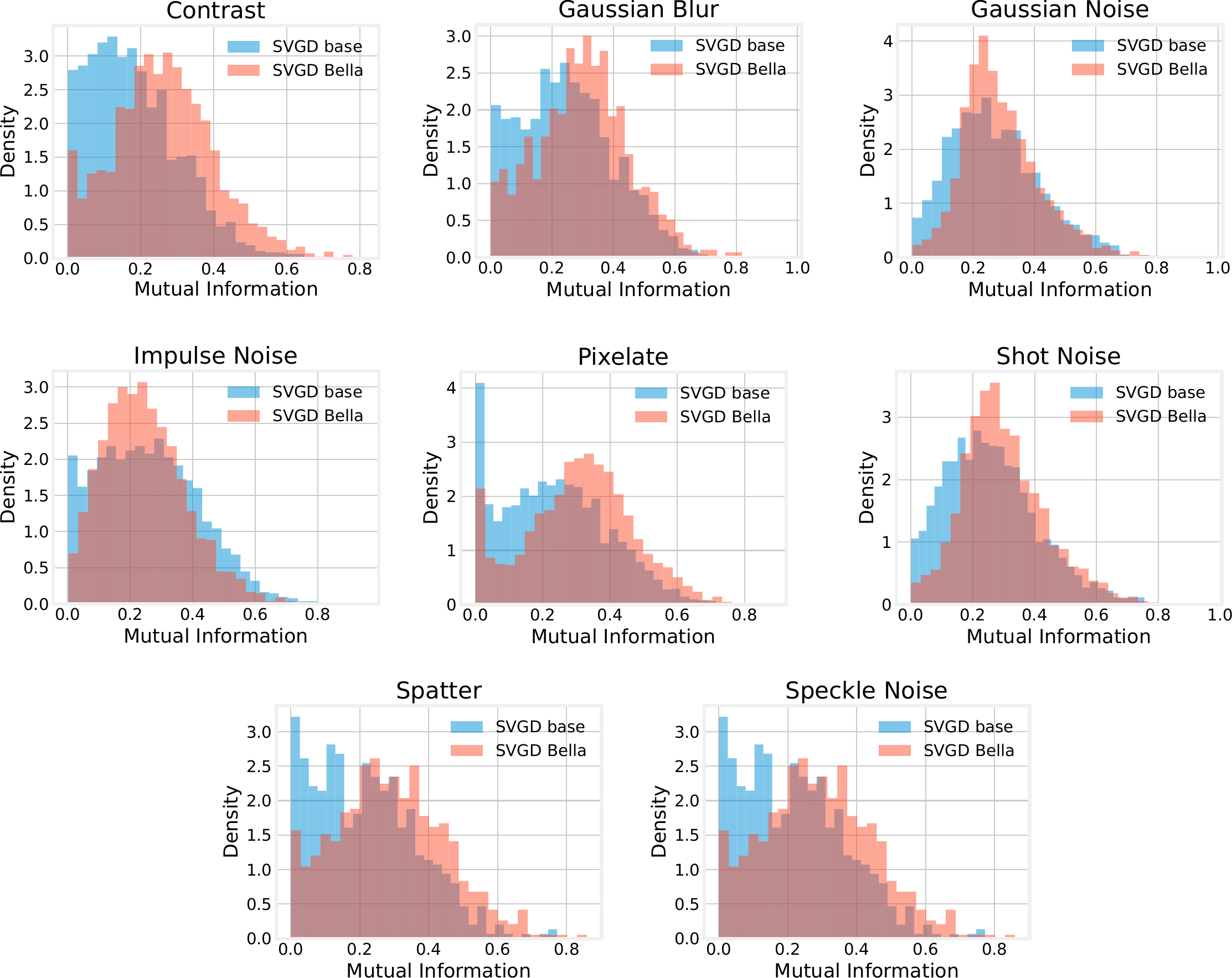}
    \caption{\small{Out-of-distribution performance of our \method method on the \texttt{CIFAR-10-C} benchmark compared with baseline SVGD models.}}
    \label{fig:ood-cifar-10-c}
\end{figure*}

\begin{table*}[!h]
\centering
\caption{Out-of-distribution performance (in terms of accuracy) of our method \method on the \texttt{CIFAR-10-C} benchmark. (Recall Bayesian approximation are from $n=5$ particles for Ensemble and SVGD).}
\setlength{\tabcolsep}{12pt} 
\resizebox{1.0\textwidth}{!}{%
\setlength{\tabcolsep}{3pt} 
\begin{tabular}{lccccccccc} 

\dtoprule
\multicolumn{1}{l}{\multirow{2}{*}{Models}} & \multicolumn{8}{c}{\texttt{CIFAR-10-C}} & \multirow{2}{*}{Average} \\ \cline{2-9}  
\multicolumn{1}{l}{}                       & Contrast & \makecell{Gaussian\\Blur} & \makecell{Gaussian\\Noise} & \makecell{Impulse\\Noise} & Pixelate & \makecell{Shot\\Noise} & Spatter & \makecell{Speckle\\Noise} & \\ \hline 
Single base                                  & 91.78    & 90.58                     & 62.96                      & 66.48                     & 77.94    & 71.56                  & 92.14   & 73.87                     & {78.53} \\ \hline

VBL                               & 83.77 & 89.34                     & 51.10                      & 52.08                     & 76.43    & 60.94                  & 89.21   & 63.19                     & {70.76} \\ \hline
Ensemble \method                               & 93.30    & 93.30                     & 64.28                      & 74.36                     & 85.86    & 75.73                  & 94.83   & 75.24                     & {82.24} \\ 
Ensemble base                                & 92.69    & 91.92                     & 72.90                      & 79.08                     & 86.49    & 78.83                  & 93.45   & 80.49                     & {84.61} \\ \hline
SVGD \method                                   & 94.69    & 94.05                     & 67.89                      & 75.86                     & 88.70    & 76.49                  & 94.85   & 77.61                     & {83.77} \\
SVGD base                                    & 97.16    & 93.02                     & 75.65                      & 81.49                     & 86.75    & 81.56                  & 95.19   & 82.89                     & 86.84 \\ \dbottomrule
\end{tabular}
}

  \label{tab:cifar10-c}
\end{table*}

\vspace{2mm}
\heading{Results.~}The data presented in Figure~\ref{fig:ood-stl} indicate that our approach slightly outperforms the SVGD and Ensemble baseline models, and significantly surpasses the capabilities of a singular network. We propose that the incorporation of a low-rank adaptation mechanism contributes to reducing overfitting, thereby enhancing the model's generalization ability on out-of-distribution (OOD) datasets. This underscores the efficacy of our method in fostering robust generalization across OOD datasets.

\section{Measure Uncertainty on Corrupted \texttt{CIFAR-10-C} Dataset}
\label{appd:mi_cifar10-c}

In this section, we apply Mutual Information (MI) as a metric for assessing uncertainty. This approach enables us to compare the performance of our \method method with models based on Stein Variational Gradient Descent (SVGD) in handling Out-of-Distribution (OOD) tasks, focusing on the CIFAR-10-C dataset. 
We adhere to the corruption types listed in \cref{tab:cifar10-c} for this analysis.

\vspace{2mm}
\heading{Results.~}The outcomes, depicted in \cref{fig:ood-cifar-10-c}, affirm the efficacy of \method. It demonstrates comparable, and in some instances, superior performance to SVGD-based models, especially notable in most cases such as involving contrast adjustments, Gaussian blur, and pixelation. This indicates a robust ability to detect increased uncertainty in OOD datasets, which is a valuable attribute for enhancing the reliability of machine learning models in unpredictable environments.

\begin{table*}[t]
\centering
\caption{Model performance comparison among `soup' models on CIFAR-10 and CIFAR-100 datasets. It should be noted that the mean of predictive entropies
has been reported for correctly and incorectly classified samples.}
\setlength{\tabcolsep}{12pt} 
\resizebox{1.0\textwidth}{!}{
\begin{tabular}{lccc|ccc}
\dtoprule
\multirow{2}{*}{Model Name} & \multicolumn{3}{c|}{\texttt{CIFAR-10}}                  & \multicolumn{3}{c}{\texttt{CIFAR-100}}                 \\ \cline{2-7} 
                            & Accuracy $\uparrow$ & \makecell{PE\\Correct $\downarrow$} & \makecell{PE\\Incorrect $\uparrow$} & Accuracy $\uparrow$ & \makecell{PE\\Correct $\downarrow$} & \makecell{PE\\Incorrect $\uparrow$} \\ \hline
Ensemble Bella soup         & 95.74    & 0.0597  & 0.6063    & 80.82    & 0.3461  & 1.2261    \\ 
Ensemble base soup          & 97.42    & 0.0986  & 0.6917    & 86.42    & 0.3025  & 1.1148    \\ \hline
SVGD Bella soup             & 96.32    & 0.0771  & 0.6988    & 83.59    & 0.2935  & 1.1044    \\
SVGD base soup              & 97.62    & 0.0438  & 0.5528    & 86.73    & 0.2609  & 1.0426    \\ \hline
Ensemble base               & 97.54    & 0.1125  & 0.5947    & 86.89    & 0.3738  & 1.0677    \\
Ensemble Bella              & 97.58    & 0.0685  & 0.5098    & 86.50    & 0.4100  & 1.1218    \\ \hline
SVGD base                   & 97.87    & 0.0572  & 0.4692    & 87.32    & 0.3256  & 1.0015    \\ 
SVGD Bella                  & 97.95    & 0.0603  & 0.5193    & 88.09    & 0.4000  & 1.0915    \\ 
\dbottomrule
\end{tabular}
}
\label{tab:soup}
\end{table*}

\section{Generalization Performance of Models on \texttt{CIFAR-10-C} Task}
\label{appd:cifar10c}
The generalization performance of different models in terms of accuracy across various noise types (with a maximum severity level of 5) is depicted in \cref{tab:cifar10-c}. The results show that Bella SVGD performs comparably to its base counterpart.

\section{Additional Comparison Studies}
\label{appd:additional}

In this section, we conduct further studies on the benefits of our approach utilizing recently developed \textit{weight averaging} concept.

\begin{figure}[!h]
  \centering

  \begin{minipage}{0.48\linewidth}
    \includegraphics[width=\columnwidth]{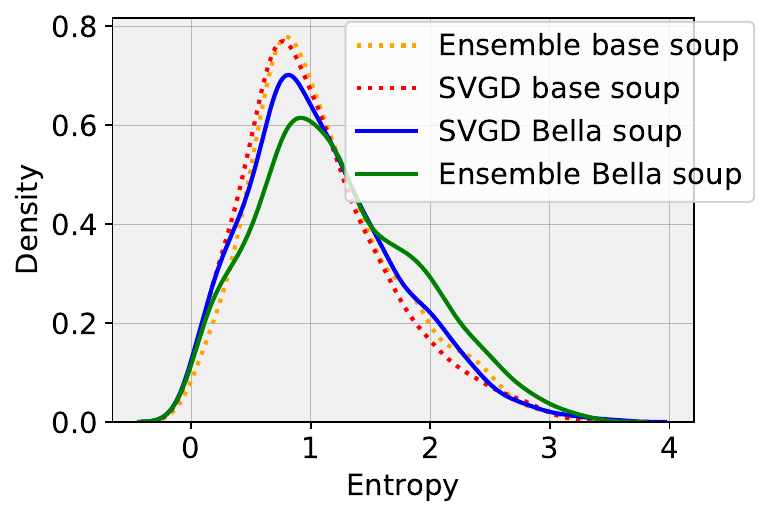}
    \caption*{(a) Miss-classified examples \texttt{CIFAR-100}}
    \label{fig:ab}
  \end{minipage}%
  \hfill
  \begin{minipage}{0.48\linewidth}
    \includegraphics[width=\columnwidth]{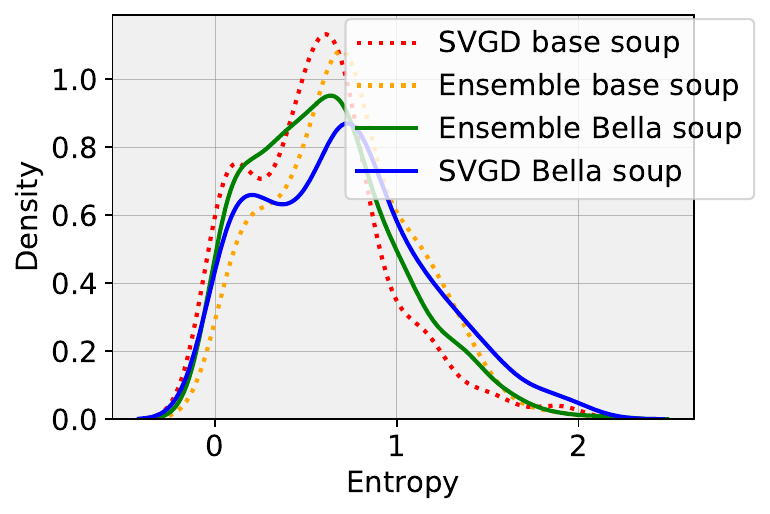}
    \caption*{(b) Miss-classified examples \texttt{CIFAR-10}}
    \label{fig:short-b}
  \end{minipage}%
  

  \begin{minipage}{0.48\linewidth}
    \includegraphics[width=\columnwidth]{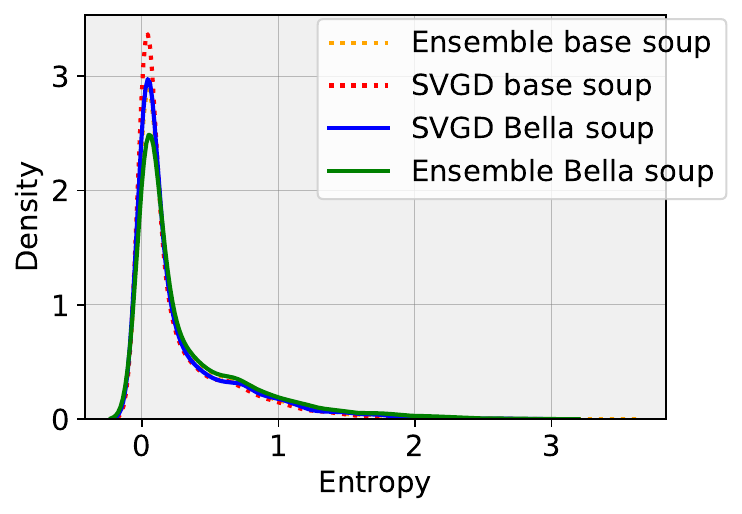}
    \caption*{(c) Correctly classified \texttt{CIFAR-100}}
    \label{fig:corrcifar100}
  \end{minipage}%
  \hfill
  \begin{minipage}{0.48\linewidth}
    \includegraphics[width=\columnwidth]{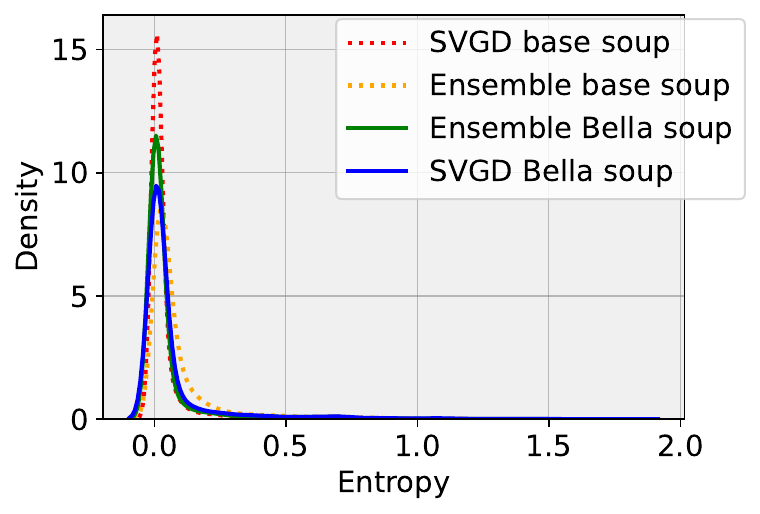}
    \caption*{(d) Correctly classified \texttt{CIFAR-10}}
    \label{fig:short-b}
  \end{minipage}%

  \caption{The density plots for predictive entropy of soup models for misclassified and correctly classified images of \texttt{CIFAR-10} and \texttt{CIFAR-100} show comparable performance of \method with baselines, despite being trained with much fewer parameters.}
  \label{fig:soup}
\end{figure}

Recent research has increasingly explored the concept of weight averaging, a technique where parameters from multiple models are combined to create a unified model that may offer superior predictive performance~\cite{matena2022merging, wortsman2022model, wortsman2022robust}. This approach is grounded in the theory that models starting from the same pre-trained state tend to have a linear loss landscape, making averaging a viable strategy. For a detailed mathematical discussion, see~\cite{rame2022diverse}. Corroborating this,~\cref{fig:soup} features density plots and~\cref{tab:soup} that illustrate the predictive entropy for samples that are correctly and incorrectly classified by the \texttt{CIFAR-10} and \texttt{CIFAR-100} datasets using the `soup' models. These results show that the performance of \method models, post-averaging, is comparable to that of more computationally intensive baseline models. This suggests that \method models, despite having fewer trainable parameters, are capable of achieving the same level of performance as their more complex counterparts.

\begin{table}[!h]
  \centering
  \caption{Choosing different layers for fine-tuning with Bella, with $\nlr=4, n=5$ on \texttt{CAMELYON17} achieves \textbf{\textit{state-of-the-art performance}}~\cite{leaderboard} (see highlighted row).
 }
\setlength{\tabcolsep}{12pt} 
\resizebox{0.60\columnwidth}{!}{
\setlength{\extrarowheight}{0pt}
\addtolength{\extrarowheight}{\aboverulesep}
\addtolength{\extrarowheight}{\belowrulesep}
\setlength{\aboverulesep}{0pt}
\setlength{\belowrulesep}{0pt}
\label{tab:ablation_layers}
\begin{tabular}{cccc} 
\toprule
\multicolumn{2}{l}{Layer name}              & \multirow{2}{*}{\begin{tabular}[c]{@{}c@{}}Trainable\\Params\end{tabular}} & \multirow{2}{*}{Accuracy}  \\ 
\cmidrule(l){1-2}
FC                                  & Proj. &                                                                            &                            \\ 
\midrule
0-11                                & 0-11  & 1.85M                                                                      & 94.89                      \\ 
\midrule
0-11                                & –     & 925K                                                                       & 95.11                      \\ 
\midrule
\rowcolor[rgb]{0.898,0.898,0.898} – & 0-11  & 925K                                                                       & \textbf{95.83}             \\ 
\midrule
0-5                                 & 0-5   & 925K                                                                       & 92.7                       \\ 
\midrule
6-11                                & 6-11  & 925K                                                                       & 91.39                      \\
\bottomrule
\end{tabular}
}%
\label{tab:ablation_layers} 
\end{table}

\begin{figure}[h]
  \centering

  \begin{minipage}{0.48\linewidth}
    \includegraphics[width=\columnwidth]{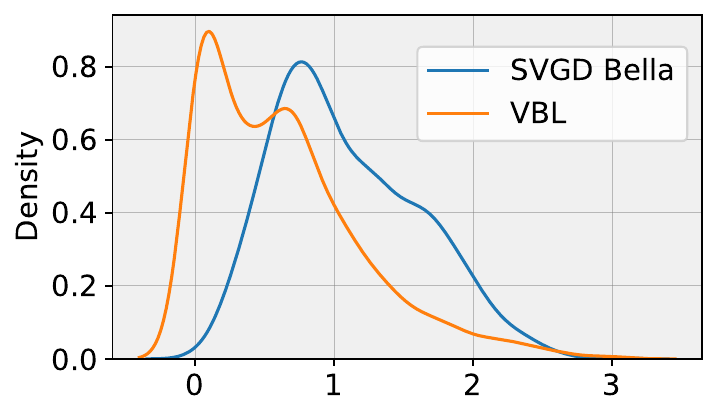}
    \caption*{(a) Miss-classified examples \texttt{CIFAR-100}}
    \label{fig:ab}
  \end{minipage}%
  \hfill
  \begin{minipage}{0.48\linewidth}
    \includegraphics[width=\columnwidth]{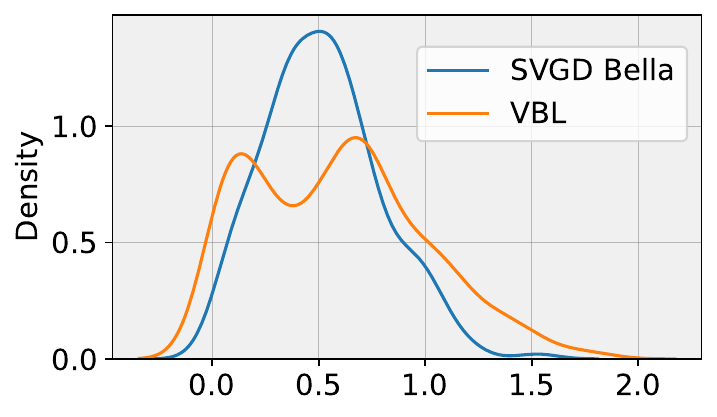}
    \caption*{(b) Miss-classified examples \texttt{CIFAR-10}}
    \label{fig:short-b}
  \end{minipage}%
  

  \begin{minipage}{0.48\linewidth}
    \includegraphics[width=\columnwidth]{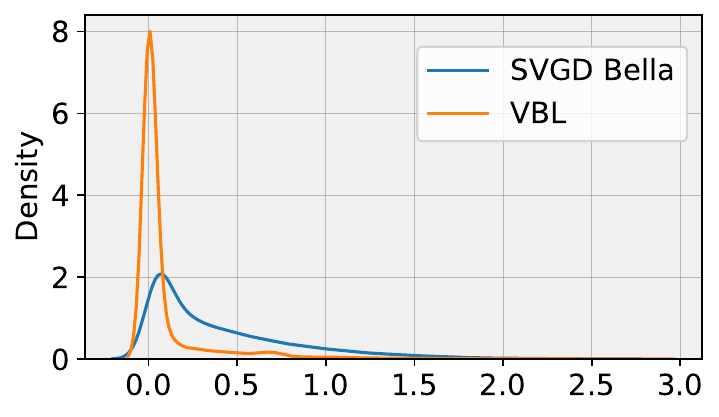}
    \caption*{(c) Correctly classified \texttt{CIFAR-100}}
    \label{fig:corrcifar100}
  \end{minipage}%
  \hfill
  \begin{minipage}{0.48\linewidth}
    \includegraphics[width=\columnwidth]{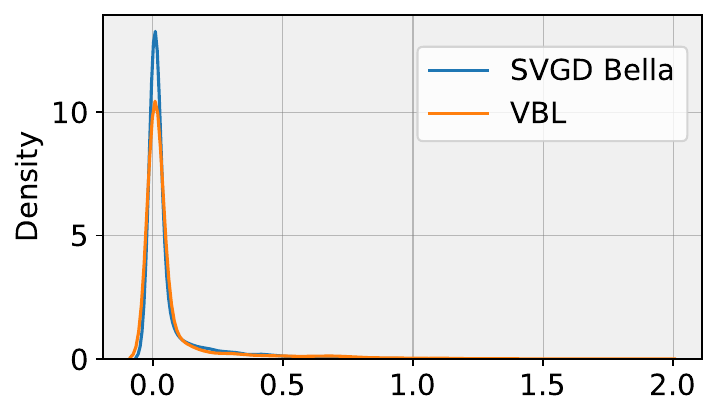}
    \caption*{(d) Correctly classified \texttt{CIFAR-10}}
    \label{fig:short-b}
  \end{minipage}%

  \caption{The density plots for predictive entropy of SVGD Bella and VBL for both misclassified and correctly classified images on \texttt{CIFAR-10} and \texttt{CIFAR-100} demonstrate that \method performs on par or even better than VBL.}
  \label{fig:vbl}
\end{figure}
\vspace{2mm}
\heading{Ablations on fine-tuned layers.~}
In our paper, we propose the concept of substituting the network's complete weight matrices with low-rank adapters to approximate BNNs. This section is dedicated to assessing the effectiveness of these low-rank adapters across various layer configurations within \textsf{CLIP}, utilizing \texttt{CAMELYON17}.  It should be noted that \textsf{CLIP} consists of 12 \textit{ResidualAttentionBlocks}, each comprising Fully Connected (FC) and Projection (Proj.) linear layers, which we aim to modify for our \method. The data, presented in~\cref{tab:ablation_layers}, \textbf{\textit{reveals an interesting performance outcome from our \method as it surpasses the SOTA in one of the experiments}}. This highlights the adaptability of our proposed approach, which does not necessitate extensive fine-tuning to select specific layers for optimal performance.


\vspace{2mm}
\heading{Comparison of SVGD Bella and VBL.~}In this section, we compare the performance of SVGD Bella and VBL in distinguishing between correctly and misclassified samples. Specifically, we assess their ability to assign high entropy to misclassified samples and low entropy to correctly classified ones. Figure~\ref{fig:vbl} shows that SVGD Bella outperforms VBL in this regard.

\section{Impact of $\gamma$ Values on Robustness.~}
\label{sec:rob-alphas}

This section assesses how the repulsive force parameter, denoted as $\gamma$, influences model robustness. Experiments were carried out on the \texttt{CIFAR-10} dataset using SVGD \method with a configuration of $r=16, n=5$, while varying $\gamma$ values. As illustrated in~\cref{tab:robustness-alphas}, adjustments in the repulsive force significantly affect robustness. Increasing $\gamma$ enhances robustness by promoting greater diversity among parameter particles. However, prioritizing diversity to an excessive degree (i.e., $\gamma$ values above 0.05) does not further enhance robustness and may even counteract improvements, although performance remains superior to scenarios with minimal repulsive force ($\gamma=0.01$).

\begin{table}[!h]
\centering
\caption{Robustness of different SVGD \method trained on different pushing parameters ($\gamma$).}
\setlength{\tabcolsep}{12pt} 
\resizebox{0.8\columnwidth}{!}{%
\begin{tabular}{lllll}
\dtoprule
                         & \multicolumn{4}{c}{Attack Budgets} \\ \hline
        \makecell{Pushing\\Parameter}                & \multicolumn{1}{c}{0.001} & \multicolumn{1}{c}{0.005} & \multicolumn{1}{c}{0.01} & \multicolumn{1}{c}{0.03} \\ \hline
$\gamma=0.01$ & \multicolumn{1}{c}{53.7}  & \multicolumn{1}{c}{50.4}  & \multicolumn{1}{c}{48.1} & \multicolumn{1}{c}{40.9} \\
$\gamma=0.025$ & 54.9    & 51.3   & 49.2   & 41.8   \\
$\gamma=0.05$  & 56.1    & 53.7   & 51.2   & 42.5   \\
$\gamma=0.1$   & 54.4    & 51.9   & 48.1   & 42     \\
$\gamma=0.2$   & 55.0    & 50.6   & 47.3   & 40.7   \\
$\gamma=0.3$   & 54.2    & 52.6   & 49.6   & 42.5  \\ \dbottomrule
\end{tabular}%
}

\label{tab:robustness-alphas}
\end{table}

\section{RBF kernel computation in SVGD}
\label{appd:rbf}
\label{sec:rbfcomputation}
We wish to evaluate RBF kernel \(k(\B_i\A_i,\B_j\A_j)\) where
\(\B_i\in \mathbb{R}^{d_{1}\times r}\),
\(\A_i\in \mathbb{R}^{r\times d_{2}}\):

{\small\begin{align*}
	&k(\B_{i}\A_{i},\B_{j}\A_{j})=\exp\left( -\frac{1}{2\sigma^2}\Vert \B_{i}\A_{i}-\B_{j}\A_{j}\Vert_{F}^2 \right)\\
	&=\exp\left(-\frac{1}{2\sigma^2}\mathrm{Tr}[(\B_{i}\A_{i}-\B_{j}\A_{j})^\top (\B_{i}\A_{i}-\B_{j}\A_{j})]\right).
\end{align*}}

 The trace term can be written as
{\small\begin{align*}
	\mathrm{Tr}[\A_{i}\A_{i}^\top \B_{i}^\top \B_{i}]&+\mathrm{Tr}[\A_{i}\A_{i}^\top \B_{i}^\top \B_{i}]\\
 &-\mathrm{Tr}[\A_{j}\A_{i}^\top \B_{i}^\top \B_{j}]-\mathrm{Tr}[\A_{i}\A_{j}^\top \B_{j}^\top \B_{i}].
\end{align*}}
To compute \(\A_j\A_i^\top\) and \(\B_j^\top \B_i\) the cost is
\(O(r^2d_2)\) and \(O(r^2d_1)\) respectively. And each trace term is of
the form
\(\mathrm{Tr}(\mathbf{U}^\top \mathbf{V})=\langle \mathbf{U},\mathbf{V} \rangle_{F}=\mathrm{vec}(\mathbf{U})^\top\mathrm{vec}(\mathbf{V})\)
which is an \(O(r^2)\) operation and requires no further matrix
multiplication.

\section{Detailed Information on the VQA Task}
\label{appd:llava}
In this section, we provide more details about the experiments we conducted on the VQA task in \cref{sec:llava}.
\subsection{Task Definition and Experiment Setup}
\textbf{Task.} 
Visual Question Answering (VQA) \cite{vqa_v1} is a free-form and open-ended task, taking as input an image and a natural-language question about the image and producing a natural-language answer as the output.
Questions in VQA require various intelligence capabilities to answer, including image recognition and object detection, as well as reasoning like commonsense reasoning.
Below is an overview of the VQA dataset utilized in our paper:
\begin{itemize}
    \item \textbf{VQA v2} \cite{vqa_v2}: This dataset contains 204,721 images, more than 1 Billion (1B) questions, and 10B ground truth answers in total.
    There are three main types of answers: Yes/No, Number, and Other. The evaluation set contains 80,541, 28,134 and 105,679 questions for Yes/No, Number, and Other respectively.
\end{itemize}
As open-ended questions may result in a diverse set of possible answers, VQA gather 10 human annotations for each question as the ground-truth answers.
These answers can be different from each other, and even incorrect.

\vspace{2mm}
\noindent\textbf{Models.} 
In our experiments, we applied \method on top of LLaVA \cite{liu2023llava,liu2023improvedllava} to address the VQA task.
Specifically, we utilized the LLaVA-1.5 7B model, an improved version of the original LLaVA with superficial modifications.
We followed the public training method using a deepspeed codebase. 

However, manipulating parameter gradients is non-trivial, and there is no public way of doing this using deepspeed.
Therefore, we simulated the low-rank approximation from the same initialization by conducting several end-to-end fine-tuning settings with different random seeds, learning rates, and gradient accumulation steps.
Together with the official public model, based on our \method, there are four generated variants we call \method-0, \method-1, \method-2, \method-3 respectively.

\noindent\textbf{Metrics.}
Below are a detailed metrics used during our evaluations.
\begin{itemize}
    \item \textbf{Accuracy}: In VQA dataset, there are 10 human annotations for each question. The model prediction accuracy is calculated by
    \begin{equation}
        Accuracy({\color{red}ans}) = \min \left\{ \frac{\# humans\ that\ said\ {\color{red}ans}}{3}, 1\right\}.
    \end{equation}
    In order to be consistent with human accuracy, this metric is averaged over all 10 chosen 9 sets of human annotations.
    
    \item \textbf{Exact Match}: We defined the metric as follows:
    \begin{equation}
        EM =\begin{cases}
                1, & \text{if $Accuracy=1$}.\\
                0, & \text{otherwise}.
            \end{cases}
    \end{equation}
    When $Accuracy$ equals 1, it means the predicted answer is 100\%, same with the ground-truth annotation, i.e., Exact Match with one another.
\end{itemize}
To measure the model uncertainty along with the human confidence in \cref{fig:correlation}, we define \textbf{Entropy} and \textbf{Human Confidence} as follows.
\begin{itemize}
    \item \textbf{Entropy}: Given a single question, we can calculate the entropy of the model prediction after applying $softmax$ over the logits of LLaVA:
    \begin{equation}
        Entropy = \frac{1}{N} \sum_{i}^{N}\sum_{j}^{V} {-p_{ij}\times \ln(p_{ij})}
    \end{equation}
    Note that $N$ is the output sequence length, $V$ is the vocabulary size, and $p_{ij}$ is the output of softmax function.
    We expect the entropy to be lower for correct predictions, as it stands for lower uncertainty.
    We expect the entropy to be higher for incorrect predictions, i.e., more uncertain on the prediction.
    
    \item \textbf{Human Confidence (HC)}: We calculate the confidence of 10 human annotators as follows:
    \begin{equation}
        HC = \frac{1}{10} \max \{\# humans\ that\ said\ {ans}\}
    \end{equation}
    More annotators agree with the same answer, higher HC will be.
\end{itemize}

\subsection{Experiments}
\heading{Correlation.~}
Given a question, our model generates a prediction accompanied by logits. We calculate the \textit{Entropy} using the logits and assess \textit{Human Confidence} by leveraging 10 ground-truth responses for this prediction. After compiling all predictions, we introduce a specific $threshold$, segregating them into two distinct categories: those not meeting the $threshold$ are classified under `Low Entropy', while those exceeding it are allocated to the `High Entropy' segment. In \cref{fig:correlation}, the $threshold$ values are set within the range \{0.1, 0.2, ..., 0.8\}. For each threshold level, we compute the average \textit{Human Confidence} for both `Low Entropy' and `High Entropy' segments, and these averages are plotted as distinct curves.

\heading{Similarity.~} 
Initially, we computed the pairwise distances among the trained \method models. We employed cosine similarity, which varies between 0 and 1, as our metric. As shown in \cref{tab:cos sim}, the choice of training strategies influences the degree of similarity among model parameters. Notably, \method-0 exhibits the greatest dissimilarity when compared to the rest, whereas \method-2 and \method-3 display the highest similarity.

\begin{table}[h]
 \caption{Performance of different LLaVA models on `Other' questions of VQA tasks.}
    \centering
    \setlength{\tabcolsep}{4pt} 
    \resizebox{\linewidth}{!}{%
    \begin{tabular}{lcccc}
    \dtoprule
        Models              & Correct Entropy & Incorrect Entropy & Accuracy & Exact Match \\ \hline
        Single base         & 0.5782          & 1.4826            & 70.00    & 57.17       \\
        \hline
        Ensemble \method    & 0.6079          & 1.5457            & 67.05    & 55.26       \\
        \hline
        SVGD \method        & 0.5359          & 1.4602            & 68.38    & 56.68       \\
    \dbottomrule
    \end{tabular}
    }  
    \label{tab:other llava}
\end{table}

\begin{table}[!h]
    \caption{Cosine Similarities between different \method trained weights.}
    \centering
    \setlength{\tabcolsep}{12pt} 
    \resizebox{0.85\columnwidth}{!}{%
    \begin{tabular}{ccccc}
    \dtoprule
                   & \method-0    & \method-1    & \method-2    & \method-3 \\ \hline
       \method-0   & 1            & 0.0286       & 0.0327       & 0.0317 \\
       \method-1   & 0.0286       & 1            & 0.6289       & 0.5508 \\
       \method-2   & 0.0327       & 0.6289       & 1            & 0.8047 \\
       \method-3   & 0.0317       & 0.5508       & 0.8047       & 1      \\
    \dbottomrule
    \end{tabular}
    }
    \label{tab:cos sim}
\end{table}

\subsection{Results}
\begin{figure*}[!t]
    \centering
    \includegraphics[width=0.9\textwidth]{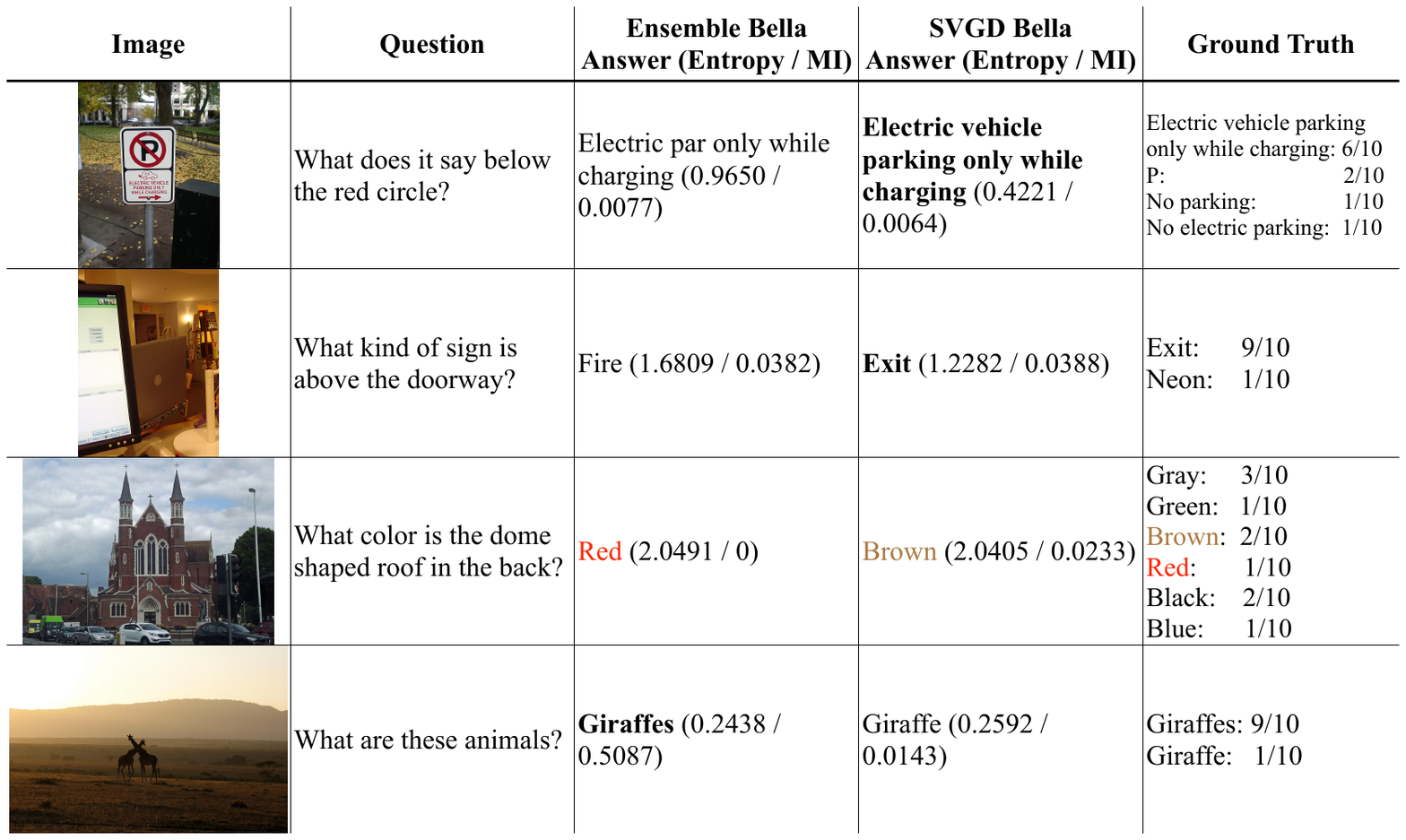}
    \caption{\small{In comparing LLaVA-VQA predictions on `Other' questions: SVGD \method outperforms Ensemble \method, even with more extended predictions. In one case, both \method err, but with notable human annotator disagreement. Interestingly, SVGD \method shows greater uncertainty (MI), unlike the confident Ensemble \method (MI=0). Given that 2 out of 10 annotators chose "Brown," SVGD \method's answer is deemed acceptable. In another case, SVGD \method correctly identifies "Giraffe" albeit singularly, mirroring a minor error also seen among human annotators.}}
    \label{fig:llava_case_study_other}
\end{figure*}

The performance of different single models and ensembles are presented in \cref{tab:vqa-llava}, \cref{tab:other llava}.
We ensembled two \method models by simply averaging two softmax output scores.
Given the model similarity matrix in \cref{tab:cos sim}, some observations and conclusions can be made from these tables.
\begin{itemize}
    \item Across all models, the averaged entropy of correct predictions is lower than the incorrect predictions. It shows models usually have higher certainty in correct predictions but lack confidence in wrong predictions.
    \item To some extent, entropy can be used as an indicator of the model's accuracy. For example, models with high accuracy often have lower entropy for correct predictions and higher entropy for incorrect predictions.
    \item We find that an ensemble of two dissimilar models, e.g., SVGD \method (\method-0, \method-1) (parameters are away from each other), usually ends with higher accuracy and exact match scores, as well as lower entropy over correct predictions. While two similar models ensemble ends with lower accuracy and exact match scores, e.g., Ensemble \method (\method-2, \method-3).
    \item For other questions in VQA, the EM score of the SVGD \method is slightly lower than the Single base. As models would generate longer textual answers for those questions, it is non-trivial to investigate a better ensemble strategy. In \cref{fig:llava_case_study_other}, we present some incorrect predictions yet reasonable generation by SVGD \method, where human annotators show high uncertainty as well.
\end{itemize}

\begin{figure*}[!t]
    \centering
    \begin{minipage}[t]{0.49\columnwidth}
        \centering
        \includegraphics[width=0.90\columnwidth]{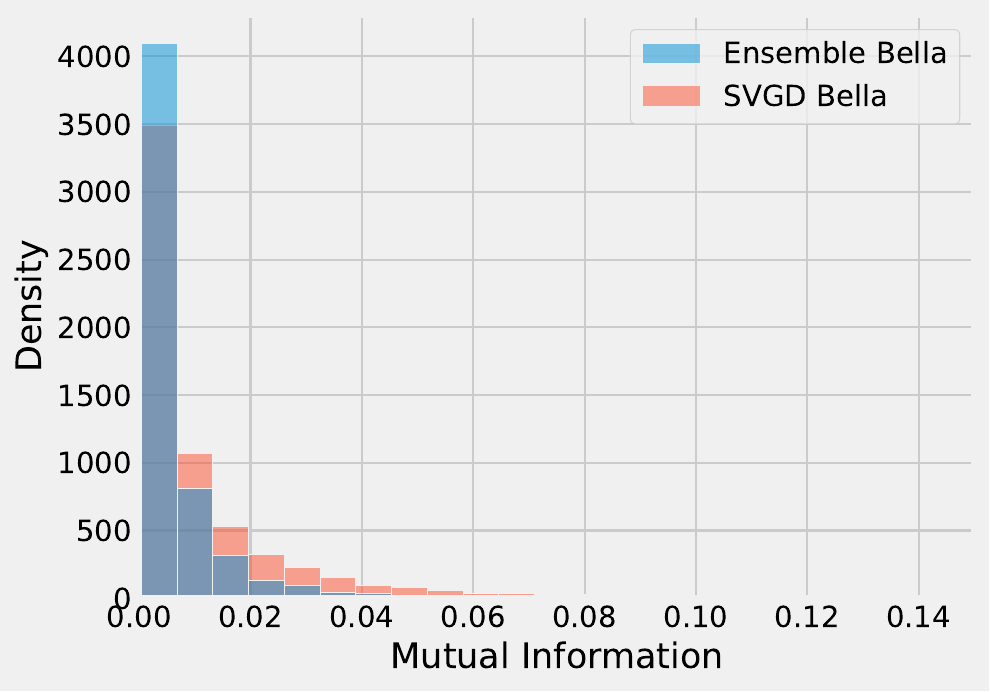}
        \caption*{(a) Yes/No questions.}
    \end{minipage}%
    \begin{minipage}[t]{0.49\columnwidth}
        \centering
        \includegraphics[width=0.95\columnwidth]{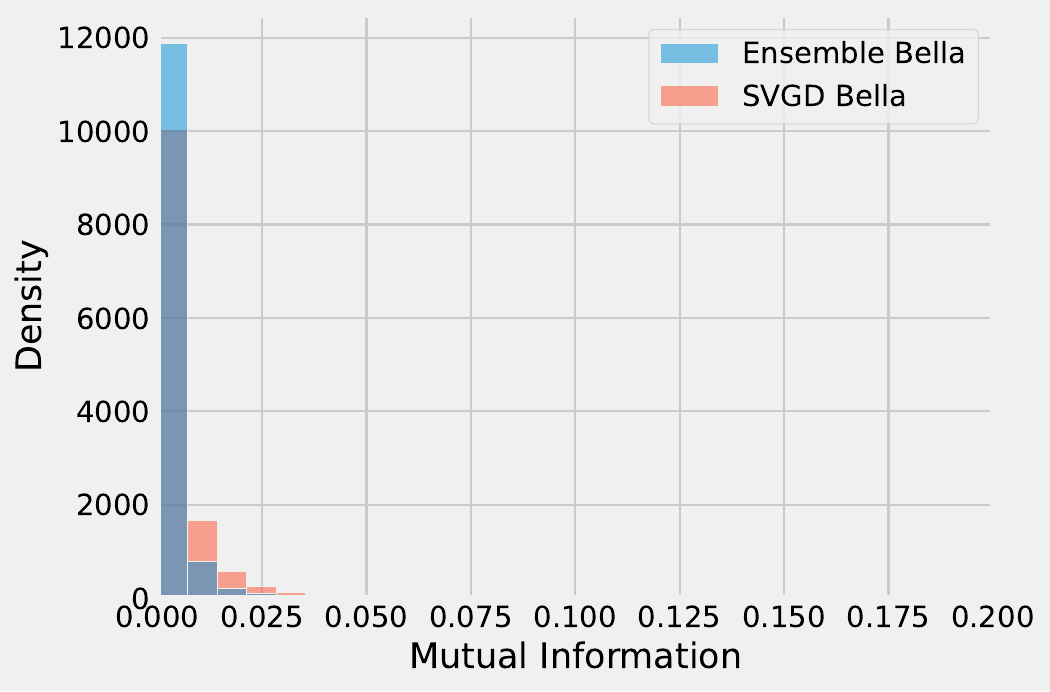}
        \caption*{(b) Number questions.}
    \end{minipage}
    \caption{\small{Evaluation of uncertainty estimations on misclassified examples using Mutual Information on VQA tasks. Overall, the SVGD \method outperforms the Ensemble \method as it shows higher uncertainty (distribution shifting $\rightarrow$).}}
    \label{fig:llava_mutual_information}
\end{figure*}

\begin{figure*}[!h]
    \centering
    \begin{minipage}[b]{0.85\textwidth}
        \includegraphics[width=0.85\columnwidth]{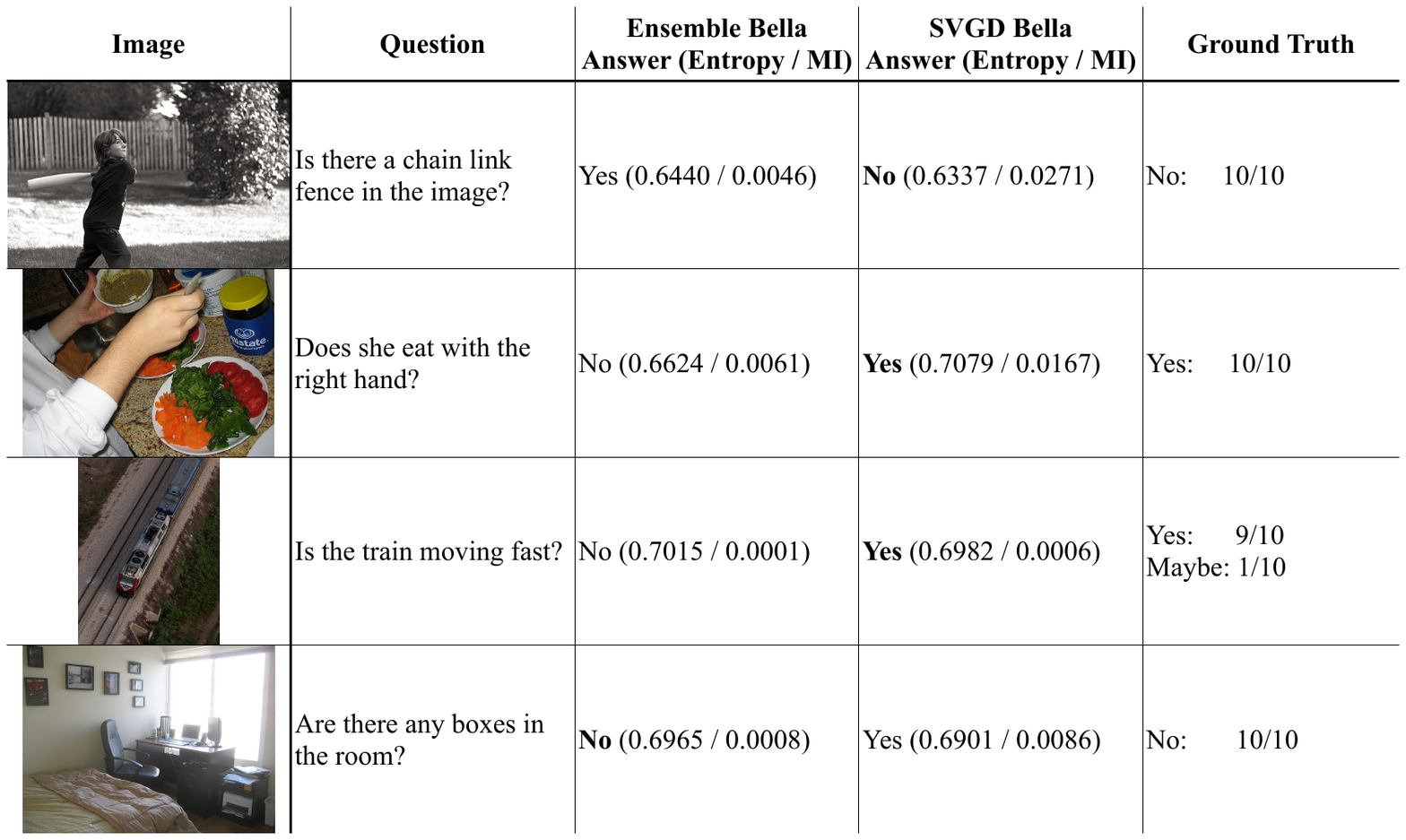}
        \caption*{\small{(a) In assessing LLaVA-VQA predictions for `Yes/No' questions, SVGD \method tends to produce accurate responses but with a notable increase in uncertainty (reflected by higher MI values). Particularly striking is the last case study, where SVGD \method exhibited a tenfold increase in uncertainty (MI) compared to the Ensemble \method, when it makes an incorrect prediction.}}
        \label{sublable1}
    \end{minipage}
    \begin{minipage}[b]{.85\textwidth}
        \includegraphics[width=0.85\columnwidth]{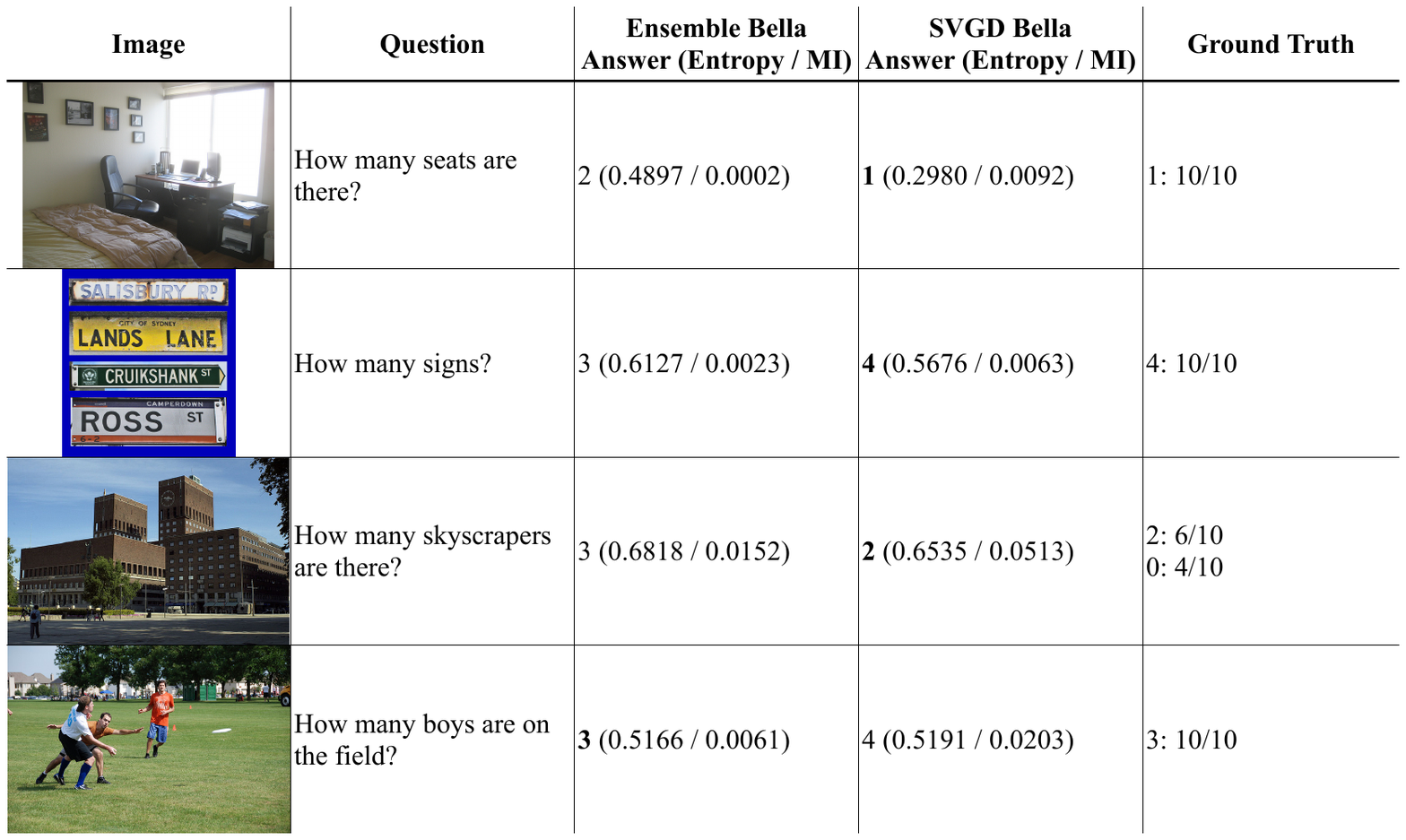}
        \caption*{\small{(b) For LLaVA-VQA predictions on `Number' questions, SVGD \method surpasses Ensemble \method in performance. A noteworthy observation from the final case study is the elevated uncertainty associated with SVGD \method during instances of incorrect predictions, indicating its lack of confidence in those particular responses. In the third case study, SVGD \method exhibited high uncertainty (MI), reflecting the varying opinions among human evaluators.}}
        \label{sublable2}
    \end{minipage}
    \caption{Model predictions and uncertainty on `Yes/No' and `Number' questions.}
    \label{fig:llava_case_study}
\end{figure*}

In addition, we show the evaluation of uncertainty using Mutual Information in \cref{fig:llava_mutual_information} and provide case studies in \cref{fig:llava_case_study_other} and \cref{fig:llava_case_study} to show predicated answers from different models.
The \cref{fig:llava_mutual_information} show the efficacy of \method. 
It enables large-scale networks like LLaVA to predict outcomes with \textit{uncertainty} for both VQA `Yes/No' and `Number' questions, a capability that a single base model lacks.

\newpage

\section{Detailed Parameters and Datasets}
\label{sec:parameters}

In this section, we mention in detail the description for each of the dataset utilized in our experiment as below:

\begin{itemize}
    \item \textbf{CIFAR-10}: This dataset comprises $60,000$ $32\times32$ color images divided into 10 distinct classes, with each class containing 6,000 images. It is partitioned into 50,000 training images and $10,000$ test images. The classes cover a range of subjects from animals to vehicles, providing a fundamental challenge in image classification.

    \item \textbf{CIFAR-10-C}: This dataset is an extension of the \texttt{CIFAR-10} dataset, designed to evaluate the robustness of machine learning models against common image corruptions. It contains the same 60,000 images as CIFAR-10, however, the images in \texttt{CIFAR-10-C} have been systematically altered using a range of corruption techniques, including noise, blur, weather, and digital effects, resulting in 19 different corruption types each at 5 severity levels. This dataset is used to test the performance of models in recognizing objects under various real-world conditions, making it a valuable tool for improving the reliability and robustness of image recognition systems.

    \item \textbf{STL-10}: This dataset is a benchmark for evaluating image recognition algorithms, featuring 13,000 color images. This dataset is divided into 5,000 training images and 8,000 test images, distributed across 10 different classes that include a variety of objects such as animals and vehicles. Each image in the dataset is $96\times96$ pixels, offering higher resolution than many similar datasets such as \texttt{CIFAR-10}. The \texttt{STL-10} dataset is tailored for supervised learning tasks in image recognition, providing a structured framework for developing and testing algorithms' ability to classify images into predefined categories.
    
    \item \textbf{CIFAR-100}: Similar to \texttt{CIFAR-10}, the \texttt{CIFAR-100} dataset is composed of $100$ classes, each with $600$ images, offering a more detailed classification challenge compared to CIFAR-10.
    
    \item \textbf{CAMELYON17}: The \texttt{CAMELYON17} dataset is utilized in a domain generalization context, where the domains are represented by different hospitals. The primary objective is to develop models capable of generalizing to data from hospitals not included in the training set. Focusing on binary classification, the dataset comprises $96\times96$ histopathological images as input, with the task to identify the presence of tumor tissue in the central $32\times32$ region, indicated by a binary label. 
    
    \item \textbf{ImageNet (ILSVRC2012)}: which is a subset of the ImageNet dataset specifically used for 
    the ImageNet Large Scale Visual Recognition Challenge in $2012$, contains over $1.2$ million images
    distributed across $1,000$ different classes.
    
    \item \textbf{DomainNet}: \texttt{DomainNet} is one of the largest and most diverse datasets available for domain adaptation studies. It contains approximately $600,000$ images across $345$ categories, spanning six distinct visual domains (Real, Clip-Art, Infograph, Paint, Sketch, Quick). This diversity in domains and categories enables the dataset to simulate real-world scenarios where models must adapt to different visual representations and styles.
\end{itemize}

\heading{Hyper-Parameters.}

Detailed information about the hyper-parameters used can be found in~\cref{tab:hyper-parameters}.
\begin{table}[t]
\centering
\caption{Hyper-parameters setting in our experiments.}
\resizebox{.7\columnwidth}{!}{%
\begin{tabular}{@{}ccc@{}}
\dtoprule
Name  & Value & Notes                                \\ \midrule
$r$     & {\texttt{ImageNet}:16, Others:4}     & Rank values                          \\
$\gamma$ &   0.01    & Weight to control the repulsive force \\
$n$ & 5 & \makecell{\#Parameter particles} \\
$optimizer$ & AdamW & With adaptive scheduler\\
\dbottomrule
\end{tabular}%
}
\label{tab:hyper-parameters}
\end{table}



\end{document}